\documentclass[12pt]{article}
\usepackage{parskip}

\usepackage{soul}
\usepackage{amssymb}
\usepackage{amsthm}
\usepackage{etoolbox}
\usepackage{lmodern}
\usepackage{amsmath,bm} 
\numberwithin{equation}{section}
\usepackage{mathtools}

\usepackage{siunitx}
\usepackage{color, soul}
\usepackage{textcomp}
\usepackage{graphicx}
\usepackage{caption}
\usepackage{indentfirst}
\usepackage{url}
\usepackage{titlesec}
\usepackage[top=1in, bottom=1.3in, left=0.875in, right=0.875in]{geometry}
\usepackage[linesnumbered,ruled]{algorithm2e}
\usepackage[titletoc,toc,title]{appendix}
\usepackage{tikz}
\usepackage{multicol}
\usepackage{multirow}
\usepackage{bbm}
\usepackage{booktabs}
\usepackage{hyperref}
\usepackage{xcolor}
\usepackage{listings}
\usepackage{array}
\usepackage{tablefootnote}
\usepackage{enumitem}
\usepackage{longtable}

\usepackage{listings}
\usepackage{xcolor}

\usepackage[
  backend=biber,
  natbib,
  citestyle=authoryear,
  style=apa,
  uniquelist=false,
  uniquename=false,
  sorting=nyt,
  maxbibnames=20,
  maxcitenames=2,
]{biblatex}
\definecolor{codegreen}{rgb}{0,0.6,0}
\definecolor{codegray}{rgb}{0.5,0.5,0.5}
\definecolor{codepurple}{rgb}{0.58,0,0.82}
\definecolor{backcolour}{rgb}{0.95,0.95,0.92}

\lstdefinestyle{mystyle}{
    backgroundcolor=\color{backcolour},   
    commentstyle=\color{codegreen},
    keywordstyle=\color{magenta},
    numberstyle=\tiny\color{codegray},
    stringstyle=\color{codepurple},
    basicstyle=\ttfamily\footnotesize,
    breakatwhitespace=false,         
    breaklines=true,                 
    captionpos=b,                    
    keepspaces=true,                 
    numbers=left,                    
    numbersep=5pt,                  
    showspaces=false,                
    showstringspaces=false,
    showtabs=false,                  
    tabsize=2
}
\usetikzlibrary{er,positioning,arrows.meta,calc,shapes}
\tikzstyle{block} = [rectangle, draw, fill=white, 
    text width=7cm, text centered, rounded corners, minimum height=3em]
\tikzset{arrow/.style = { thick,  ->, >=Triangle}}

\usepackage[labelformat=simple]{subcaption}
\newcolumntype{P}[1]{>{\centering\arraybackslash}p{#1}}

\def\spacingset#1{\renewcommand{\baselinestretch}{#1}\small\normalsize} 
\spacingset{1}
\SetKwComment{Comment}{/* }{ */}

\title{Claim Automation using Large Language Model}
\author{Zhengda Mo\thanks{Actuarial and Risk Management Sciences, University of Illinois at Urbana-Champaign, zhengda2@illinois.edu} \and Zhiyu Quan\thanks{Actuarial and Risk Management Sciences, University of Illinois at Urbana-Champaign, zquan@illinois.edu} \and Eli O'Donohue\thanks{PCMI Corporation, eli.odonohue@pcmicorp.com} \and Kaiwen Zhong\thanks{Actuarial and Risk Management Sciences, University of Illinois at Urbana-Champaign, kaiwen4@illinois.edu}}

\date{}

\begin{document}
\setstcolor{red}
\maketitle

\begin{abstract}

While Large Language Models (LLMs) have achieved strong performance on general-purpose language tasks, their deployment in regulated and data-sensitive domains, including insurance, remains limited. Leveraging millions of historical warranty claims, we propose a locally deployed governance-aware language modeling component that generates structured corrective-action recommendations from unstructured claim narratives. We fine-tune pretrained LLMs using Low-Rank Adaptation (LoRA), scoping the model to an initial decision module within the claim processing pipeline to speed up claim adjusters' decisions. We assess this module using a multi-dimensional evaluation framework that combines automated semantic similarity metrics with human evaluation, enabling a rigorous examination of both practical utility and predictive accuracy. Our results show that domain-specific fine-tuning substantially outperforms commercial general-purpose and prompt-based LLMs, with approximately 80\% of the evaluated cases achieving near-identical matches to ground-truth corrective actions. Overall, this study provides both theoretical and empirical evidence to prove that domain-adaptive fine-tuning can align model output distributions more closely with real-world operational data, demonstrating its promise as a reliable and governable building block for insurance applications.

\vspace{0.75cm}
\noindent \textbf{Keywords:} Large language model; fine-tuning; warranty claim automation; 
\end{abstract}

\newpage

\section{Introduction}

Actuarial practice has historically relied almost exclusively on structured data. For decades, risk factors derived from policyholder and development factors observed throughout the policy life cycle have constituted the core, though fundamentally limited, inputs for actuarial modeling. Yet, the ground truth of risk often resides in narrative text, such as claim descriptions, physician diagnostic notes. These narratives embed causal reasoning, contextual nuance, and operational intent that structured fields cannot capture. Ignoring this textual layer means modeling only a partial and often distorted view of the underlying risk.

Early attempts to integrate textual information using traditional Natural Language Processing (NLP) into actuarial models made only limited progress. 

\noindent\textbf{The Text-to-Feature Paradigm} The traditional approach to incorporating text into actuarial models has been to convert unstructured text narratives into manually engineered or compressed numerical features. A series of representative studies illustrates how this feature-engineering approach has remained fundamentally unchanged even though it has become more sophisticated. \citet{FrancisFlynn2010}, for example, use commercial text-mining software to extract keywords and thematic categories from claim description narratives, adding these as categorical features into GLM-based models. 
\citet{Lee2020} project short claim descriptions into a word-embedding space and construct continuous semantic risk factors by measuring similarities between claims and predefined risk-related concept words, which are then incorporated into traditional GLM/GAM frameworks for claim classification and large claim loss analysis. 
\citet{Liao2020} apply topic modeling and sentiment analysis to non-claim call transcripts to convert customer communications into service topics and sentiment indicators, framing insurance call text primarily as a source of engineered features for operational analytics. 
\citet{manski_extracting_2021} propose an interpretable text-to-loss modeling framework that transforms claim descriptions into high-dimensional word-similarity features via embeddings, and applies a group-lasso-regularized generalized additive model to predict claim severity, achieving automatic word selection and parsimonious loss prediction from unstructured text. 
\citet{zappa_text_2021} expand this idea by applying a full text-mining pipeline to NHTSA accident narratives and converting selected linguistic patterns into keyword-based indicators for use as features in traditional ratemaking models. 
With the advent of modern NLP, \citet{xu_bert-based_2022} apply BERT-based contextual representations to warranty claim texts and incorporate them into neural network models for claim grouping and severity prediction, showing that deep language embeddings substantially enhance both accuracy and stability in warranty loss modeling. 
\citet{shiNonLifeInsuranceRisk2023} demonstrate the use of categorical embeddings to transform high-dimensional categorical features into numerical vector representation, enabling more efficient information extraction from high-cardinality features. 
\citet{xu2023framework} use BERT to embed claim description narratives into fixed-length vectors and feed these embeddings into neural-network models for frequency and severity prediction—an advance in representation but still firmly within the text-to-feature paradigm. 
More recently, \citet{DongQuan2025} provide a practice-oriented survey of NLP applications in insurance, demonstrating how alternative textual data from an InsurTech partner can be transformed into de-biased and compressed risk representations that enrich traditional rating factors and enable novel industry classification for actuarial analysis. 
\citet{Zhang11112024} propose an NLP-driven framework for automated retrieval, summarization, and clustering of cyber risk literature, introducing CyLit as a domain-specific repository that supports semantic exploration and trend analysis of a rapidly evolving research field. 
\citet{cao2024assessing} leverage large-scale judgment texts and state-of-the-art (SoTA) pretrained language models to predict litigation outcomes and apply explainable AI techniques to identify influential legal reasoning, highlighting the role of machine learning in quantifying and managing litigation risk in insurance disputes. 
\citet{gan2025using} explore an imputation method based on generative adversarial networks (GANs).
Despite the growing technical sophistication of these pipelines, they all share the same limitation: linguistic content is ultimately compressed into engineered numerical features, leaving no room for text-to-text reasoning or direct modeling of linguistic structure.


While this `text-as-auxiliary-signal' strategy can produce modest improvements compared to simply ignoring unstructured data, it comes at a substantial cost: most of the contextual dependencies, semantic interactions, and pragmatic cues that give language its meaning are stripped away in the process. The emergence of Large Language Models (LLMs) offers a path beyond feature engineering, enabling actuarial models to gain insights directly from raw text. However, their off-the-shelf, API-based deployment introduces new challenges, including domain misalignment, opaque inference, and even hallucination. It also includes governance complexities that limit their suitability for regulated actuarial workflows. 

\noindent\textbf{The General-Purpose LLM Paradigm} A newer line of research investigates whether general-purpose foundation models can operate directly on insurance narratives. Representative studies illustrate this emerging paradigm. 
\citet{Balona2024ActuaryGPT} demonstrate how models such as GPT-4 can be applied across a wide range of actuarial and insurance tasks—including claims triage, fraud detection, compliance assessment, and underwriting—highlighting the shift toward end-to-end, context-aware reasoning over textual data. 
\citet{liTextualAnalysisInsurance2025} develop an API-driven LLM framework that integrates OpenAI's embedding and prompting interfaces to detect and analyze semantic discrepancies in insurance and risk-management texts, combining embedding-based distance metrics with prompt-based relational reasoning to achieve scalable, robust discrepancy assessment beyond traditional NLP pipelines. 
\citet{Hatzesberger2025GenAIActuarial} further broaden the landscape through four implemented case studies that integrate generative AI into text, document, image, and multi-agent workflows, applying foundation models to tasks such as claim prediction, regulatory data extraction, car damage classification, and autonomous reporting. 
Despite the breadth of these investigations, they share two structural limitations: the reliance on general-purpose models pretrained on largely undisclosed, domain-misaligned corpora, and the dependence on externally hosted cloud APIs that raise significant concerns around data privacy and regulatory compliance. In practice, some organizations have already embedded commercial general-purpose LLMs into operational workflows despite having limited visibility into what these models actually generate or how their outputs are systematically shaped.\footnote{\url{https://openai.com/index/introducing-openai-frontier/}}

In this work, we address these limitations by proposing a domain-adapted LLM workflow for claim automation that is explicitly designed to operate under actuarial model-governance constraints. We introduce a locally deployed framework that supports practical claim handling decisions while maintaining control, transparency, and auditability. The system focuses on a narrowly scoped intermediate automation task, generating structured corrective-action summaries based on historical claim information, allowing language understanding to be integrated into existing claim workflows. By explicitly separating linguistic interpretation from downstream financial modeling, the framework is designed to overcome both the compression limits of text-to-feature pipelines and the model governance challenges of externally hosted general-purpose LLMs systems.

Empirically, we provide direct evidence that domain-specific fine-tuning fundamentally reshapes both output formality and semantic behavior. Using large-scale historical warranty claims, we show that the fine-tuned model moves beyond generic or loosely plausible responses and instead produces corrective actions whose format, category structure, and semantic patterns closely match observed claim-handling practices. This distributional realignment places the model in a stable, operationally interpretable regime, where generated outputs can be meaningfully compared with real-world claim processes and treated as actuarially relevant signals rather than isolated predictions.

To support this analysis, we develop an actuarial model governance framework that decomposes model behavior along complementary dimensions of structural validity, semantic alignment, and distributional consistency. Together, these criteria enable systematic assessment of whether the generated outputs satisfy operational constraints, reflect correct claim semantics, and reproduce empirical claim patterns beyond observational-level accuracy. Collectively, these results establish both the empirical basis and the methodological tools required to rigorously evaluate the reliability of language-based components within claim automation workflows.

\section{Automation Framework}

\subsection{Framework Overview}

We develop a novel automation framework designed to overcome the limitations of both preceding paradigms by integrating an open-source LLM into the warranty claim process with a focus on transparency, domain alignment, and operational feasibility. Its architecture is built on four core design principles with practicality in mind:

\noindent\textbf{Local Deployment and Governance Control} Rather than relying on external API-based models, we deploy a DeepSeek-R1-Distill-Llama-8B model \citep{deepseek2025r1}, a distilled reasoning-focused variant of the DeepSeek-R1 family built on an 8B-parameter LLaMA backbone (hereafter DeepSeek-R1), on-premises and in a controlled computing environment. This ensures that all sensitive claim information remains within the organizational perimeter. It also enables full reproducibility of model outputs and provides the transparency required under stringent regulatory governance and model risk management standards.

\noindent\textbf{Domain Adaptation Fine-Tuning} We fine-tune the foundation model using the company's internal historical warranty claim information \citep{hu2022lora}. This process aligns the model's internal representations with the specific linguistic patterns, technical terminology, and action taxonomies that characterize real-world claim management. Such representation alignment has been shown to improve generalization under distributional shift in cross-domain learning settings \citep{chen2024}. The resulting system reflects the insurer's current data-generating process, creating a stark contrast with the distributional mismatch of general-purpose models.

\noindent\textbf{Modular Intermediate Task Design} Instead of an end-to-end architecture that directly outputs financial predictions—a black-box approach fraught with validation challenges—we assign the model a single, well-scoped intermediate task: generating a structured corrective-action output from unstructured complaint and cause narratives. This design mirrors the natural decision sequence in the claim lifecycle, supports claim adjusters in making efficient approvals, and isolates the language-understanding component from downstream actuarial modeling. The task remains stable, easier to supervise, and consistent with established claim management practices.

\noindent\textbf{Multi-Dimensional Evaluation Strategy} We evaluate the system using a combination of semantic similarity measures and structured-output validation checks, moving beyond simple accuracy to provide a holistic assessment of the model's performance and reliability in a real-world production setting. The human-in-the-loop approach derives a gold-standard evaluation metric and provides a path for active learning and continuous improvement. 

By uniting four core design principles, our framework offers a principled, practical path for incorporating LLMs into daily claim management workflows.

\subsection{Mathematical Formulation}
\label{sec:formulation}
\maketitle

\subsubsection{Objective Function}

We formulate the task as a conditional language generation problem. Let \( \boldsymbol{x} = (x_1, x_2, \ldots, x_n) \) denote the input token sequence. For the warranty correction prediction task, the input token sequence includes a customer complaint and the associated cause of the complaint. The goal of the conditional language generation is to generate the following output sequence, \( \boldsymbol{y} = (y_1, y_2, \ldots, y_T) \), conditioned on \( \boldsymbol{x} \), in claim automation, the correction sequence actions for the product under warranty.

Specifically, the conditional language generation can be formulated as
\begin{equation}
P(\boldsymbol{y} \mid \boldsymbol{x}; \boldsymbol{\theta}) = \prod_{t=1}^{T} P(y_t \mid \boldsymbol{y}_{<t}, \boldsymbol{x}; \boldsymbol{\theta}),
\label{eq:conditional}
\end{equation} 
where \( \boldsymbol{y}_{<t} = (y_1, \ldots, y_{t-1}) \) denotes the prefix of previously generated tokens, and \( \boldsymbol{\theta} \) denotes the model parameters. To model conditional probabilities, we have several methodological options, each leading to distinct model architectures and learning frameworks, see \citet{DongQuan2025} and references therein.

Then, the model is trained by minimizing the negative log-likelihood over an observed dataset \( \mathcal{D} = \{ (\boldsymbol{x}^{(i)}, \boldsymbol{y}^{(i)}) \}_{i=1}^N \):
\begin{equation}
\mathcal{L}_{\text{MLE}}(\boldsymbol{\theta}) = - \sum_{i=1}^{N} \sum_{t=1}^{T^{(i)}} \log P\left( y^{(i)}_t \mid \boldsymbol{y}^{(i)}_{<t}, \boldsymbol{x}^{(i)}; \boldsymbol{\theta} \right),
\end{equation}
where \( T^{(i)} \) denotes the length of the \( i \)-th target sequence.

\subsubsection{Overview of Model Architecture}

\begin{figure}[h!]
    \centering
    \includegraphics[width=1\textwidth]{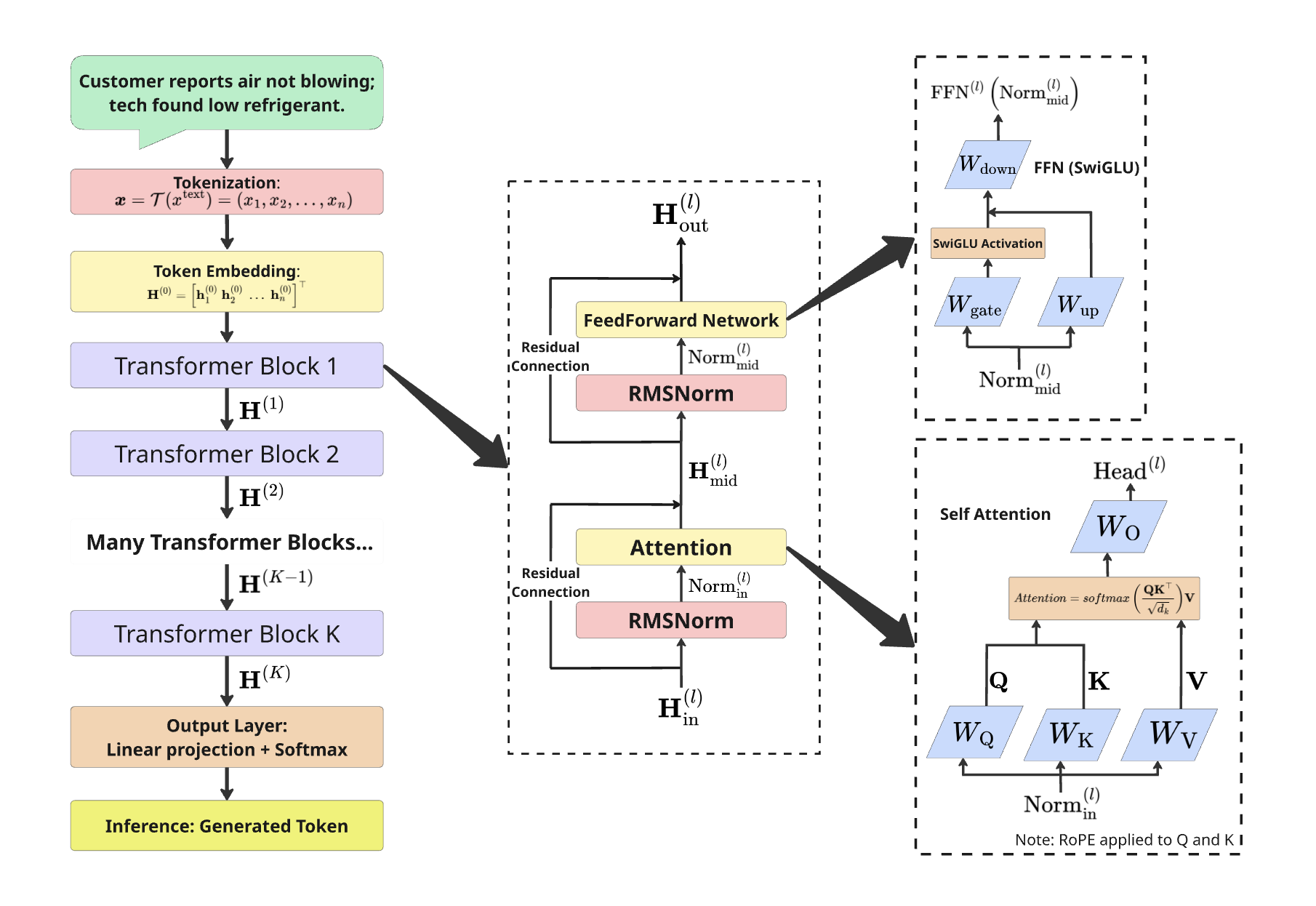}
    \caption{Overview of the token-level generation architecture used for claim automation.}
    \label{fig:model-architecture}
\end{figure}

Figure~\ref{fig:model-architecture} provides a high-level overview of our autoregressive Transformer model for claim automation. Given a customer complaint in natural language, i.e., ``The car's air conditioning is not working'', the input is tokenized and embedded into a sequence of vectors, which is then processed by a stack of Transformer decoder blocks. At each position, the final hidden state is projected into the vocabulary space via a linear layer and softmax to predict the next probable token in an autoregressive manner.

In the following sections, we formalize each component of this architecture and present its mathematical formulation.

\subsubsection{Input Representation}
\paragraph{Tokenization}

Each input \( \boldsymbol{x} \) is obtained by applying a tokenizer to a raw text string. Let \( x^{\text{text}} \in \mathcal{U}^* \) denote such a raw input string composed of Unicode characters encoded in \textit{UTF-8}. We can use the tokenizer provided with the selected pre-trained model. For example, \textit{DeepSeek-R1-Distill-Llama-8B} uses the \textit{SentencePiece} tokenizer. Specifically, the tokenizer 
\[
\mathcal{T}: \mathcal{U}^* \rightarrow \mathcal{V}^*,
\]
where \( \mathcal{V} \) is a predefined subword vocabulary. The tokenizer $\mathcal{T}$ segments the raw input string into subword tokens from \( \mathcal{V} \). 
We define the resulting token sequence as
\begin{equation}
\boldsymbol{x} = \mathcal{T}(x^{\text{text}}) = (x_1, x_2, \ldots, x_n), \quad x_s \in \mathcal{V}, \quad s=1, \dots, n.
\end{equation}

Throughout this work, we use \( \boldsymbol{x} \) to denote the tokenized model input derived from raw text.

\paragraph{Token Embedding}

Let $\boldsymbol{x} = (x_1, x_2, \dots, x_n) \in \mathcal{V}^*$ denote the tokenized model input sequence. Each token $x_s$ is mapped to a vector representation via a token embedding matrix $E \in \mathbb{R}^{|\mathcal{V}| \times d}$, producing:
\[
\bm{h}^{(0)}_s = E[x_s] \in \mathbb{R}^d,
\]
where $d$ is the embedding dimension. In this work, we adopt the pretrained token embedding matrix provided by the backbone Transformer, which is optimized end-to-end together with all model parameters during large-scale pretraining. 
The superscript $(0)$ indicates the initial input before the first Transformer block.

The token embeddings are assembled to form the input matrix:
\[
\mathbf{H}^{(0)} =
\begin{bmatrix}
(\mathbf{h}_1^{(0)})^\top \\
(\mathbf{h}_2^{(0)})^\top \\
\vdots \\
(\mathbf{h}_n^{(0)})^\top
\end{bmatrix}
\in \mathbb{R}^{n \times d},
\]
where each row of \( \mathbf{H}^{(0)} \) represents the embedding of a token in the sequence.


Throughout the following sections, we will consistently use \( \mathbf{h} \) to denote the vector representation of a single token, and \( \mathbf{H} \) to denote the sequence matrix representation formed by stacking all token vector representations.

\paragraph{Rotary Position Embedding (RoPE)}

Since the Transformer model lacks a built-in mechanism for capturing word order, it is necessary to introduce positional information externally. Rather than simply adding absolute or relative positional encodings, we adopt the Rotary Position Embedding (RoPE) method ~\citep{su2021roformer}, which integrates positional information directly into the attention mechanism in a more seamless and rotation-invariant manner. RoPE introduces relative positional information through complex-valued rotation applied to token embeddings.

Let $\bm{h}_s^{(0)} \in \mathbb{R}^d$ denote the input embedding vector for token $x_s$. We reshape it into $d/2$ two-dimensional components and interpret them as complex numbers to construct a complex-valued vector:
\begin{equation}
\tilde{\bm{h}}_s^{(0)} = \left(h^{\mathrm{re}}_{s,1} + i h^{\mathrm{im}}_{s,1},\ \ldots,\ h^{\mathrm{re}}_{s,d/2} + i h^{\mathrm{im}}_{s,d/2}\right),
\end{equation}

where each pair $(h^{\mathrm{re}}_{s,k}, h^{\mathrm{im}}_{s,k})$, for $k = 1, \ldots, d/2$, corresponds to the $(2k{-}1)$-th and $2k$-th dimensions of $\bm{h}_s^{(0)}$. The superscripts $re$ and $im$ indicate the real and imaginary parts of the constructed complex-valued two-dimensional components, respectively. To implement RoPE, we define the angular frequency vector \( \boldsymbol{\vartheta} \) using exponentially scaled values:
\begin{equation}
\boldsymbol{\vartheta}_k = \omega^k, \quad \omega = 10000^{-2/d}, \quad k = 1, 2, \dots, d/2
\end{equation}

Then, for token position \( s \), \textit{RoPE} applies a position-dependent complex-valued rotation to each component:
\begin{equation}
\bm{\tilde{h}}_{s,k}^{(0,\text{rot})} = \bm{\tilde{h}}_{s,k}^{(0)} \cdot e^{i \vartheta_k s}
\end{equation}

The superscript $rot$ indicates that the vector has been rotated. The rotated vector is then converted back to a real-valued vector \( \boldsymbol{h}_s^{(0, \text{RoPE})} \in \mathbb{R}^d \) by interleaving real and imaginary parts. This operation is applied elementwise to the query and key embeddings in self-attention:

\begin{equation}
Q' = \text{RoPE}(Q), \quad K' = \text{RoPE}(K).
\end{equation}

Here, \( Q \) and \( K \) denote the query and key matrices which will be formally introduced in Subsection~\ref{sec:multihead-attn}.

Although the formulation involves complex arithmetic, \textit{RoPE} can be efficiently implemented in real space using interleaved sine and cosine rotations for each 2D subspace and create a block diagonal rotation matrix for higher dimensions. This allows relative positional encoding while maintaining compatibility with standard Transformer architectures.

\subsubsection{Transformer Block}

We describe the Transformer ~\citep{vaswani2017attention} block structure using \textit{DeepSeek-R1} ~\citep{deepseek2025r1}, which is one of the successful decoder-only language models. Each block consists of two sub-layers: a \textit{multi-head causal self-attention module} and a \textit{feedforward network} (\textit{FFN}), both equipped with pre-normalization and residual connections.

\paragraph{PreNorm Architecture with RMSNorm}

Let $\boldsymbol{h}^{(l)}_{\text{in}} \in \mathbb{R}^d$ denote the input to the $l$-th Transformer block. \textit{DeepSeek-R1} adopts a \textit{PreNorm} architecture, where each sub-layer---either the self-attention or FFN---is preceded by normalization and followed by a residual connection:

\begin{equation}
\boldsymbol{h}^{(l)}_{\text{out}} = \boldsymbol{h}^{(l)}_{\text{in}} + \mathcal{F}^{(l)}\left( \text{Norm}(\boldsymbol{h}^{(l)}_{\text{in}}) \right),
\label{eq:prenorm}
\end{equation}
where $\mathcal{F}^{(l)}(\cdot)$ denotes the functional transformation of the $l$-th block (i.e., \textit{attention} or \textit{FFN}). For the normalization function, $\text{Norm}(\cdot)$, \textit{DeepSeek-R1} uses \textit{Root Mean Square Layer Normalization} (RMSNorm) instead of the standard LayerNorm where it normalizes the inputs of each layer across input vector for each individual data sample. Specifically, for an input vector $\boldsymbol{v} = (v_1, \dots, v_d) \in \mathbb{R}^d$, \textit{RMSNorm} ~\citep{zhang2019rmsnorm} is defined as:

\begin{equation}
\text{RMSNorm}(\boldsymbol{v}) = \frac{\boldsymbol{v}}{\sqrt{\frac{1}{d} \sum_{i=1}^d v_i^2 + \epsilon}} \cdot \boldsymbol{\gamma},
\label{eq:rmsnorm}
\end{equation}
where $\boldsymbol{\gamma} \in \mathbb{R}^d$ is a learned scaling parameter and $\epsilon$ is a small constant for numerical stability. As shown in the definition, \textit{RMSNorm} does not subtract the mean of the input vector, making it more efficient than LayerNorm. In \textit{DeepSeek-R1}, it is applied before both the attention and FFN modules. This \textit{PreNorm} + \textit{RMSNorm} design enhances training stability in deep decoder-only architectures.

\paragraph{Multi-Head Causal Self-Attention}
\label{sec:multihead-attn}

The first sub-layer in each Transformer block is a \textit{multi-head causal self-attention module}. Given an input hidden state matrix $\boldsymbol{H}^{(l)}_{\text{in}} \in \mathbb{R}^{n \times d}$, where $n$ is the sequence length and $d$ is the hidden dimension, the input is first normalized via \textit{RMSNorm}:
\begin{equation}
    \text{Norm}^{(l)}_{\text{in}} = \text{RMSNorm}(\boldsymbol{H}^{(l)}_{\text{in}}).
    \label{eq:rmsnorm_input}
\end{equation}

Each attention head $i = 1, \ldots, m_{head}$ computes the query, key, and value matrices using learned projections:
\begin{equation}
    \boldsymbol{Q}_i = \text{Norm}^{(l)} W_{i, \text{Q}}^{(l)}, \quad
    \boldsymbol{K}_i = \text{Norm}^{(l)} W_{i, \text{K}}^{(l)}, \quad
    \boldsymbol{V}_i = \text{Norm}^{(l)} W_{i, \text{V}}^{(l)},
    \label{eq:attention_projections}
\end{equation}
where $W_{i, \text{Q}}^{(l)}, W_{i, \text{K}}^{(l)}, W_{i, \text{V}}^{(l)} \in \mathbb{R}^{d \times d_k}$ are head-specific projection matrices for layer $l$, and $d_k = d / m_{head}$.

To enforce autoregressive behavior, a causal attention mask $M \in \mathbb{R}^{n \times n}$ is applied:
\begin{equation}
    M_{i,j} =
    \begin{cases}
        0 & \text{if } j \leq i \\
        -\infty & \text{if } j > i
    \end{cases}
    \label{eq:causal_mask}
\end{equation}

Each head then computes masked scaled dot-product attention ~\citep{vaswani2017attention}:
\begin{equation}
    \text{head}_i = \text{softmax} \left( \frac{\boldsymbol{Q}_i \boldsymbol{K}_i^\top}{\sqrt{d_k}} + M \right) \boldsymbol{V}_i.
    \label{eq:attention_head}
\end{equation}

The outputs of all heads are concatenated and projected back to the original hidden dimension:
\begin{equation}
    \text{MultiHead}^{(l)} = \text{Concat}(\text{head}_1, \ldots, \text{head}_{m_{head}}) W^O,
    \label{eq:multihead_concat}
\end{equation}
where $W^O \in \mathbb{R}^{d \times d}$ is a learned output projection matrix.

Finally, a \textit{residual connection} is applied to produce the output of the attention sub-layer:
\begin{equation}
    \boldsymbol{H}^{(l)}_{\text{mid}} = \boldsymbol{H}^{(l)}_{\text{in}} + \text{MultiHead}^{(l)}.
    \label{eq:attention_output}
\end{equation}

\paragraph{Feedforward Network}

The second sub-layer in each Transformer block is a position-wise FFN, applied independently to each token. Like the attention sub-layer, it uses \textit{RMSNorm} followed by a residual connection:

\begin{equation}
    \text{Norm}^{(l)}_{\text{mid}} = \text{RMSNorm}(\boldsymbol{H}^{(l)}_{\text{mid}}),
    \label{eq:ffn_norm}
\end{equation}

Each token position is transformed using two learned linear projections separated by nonlinearity. In \textit{DeepSeek-R1}, the nonlinearity is implemented via the \textit{SwiGLU}~\citep{shazeer2020glu} activation. The FFN is defined as:

\begin{equation}
    \text{FFN}^{(l)}(x) = \text{SwiGLU} \left( x W_{\text{up}}^{(l)},\; x W_{\text{gate}}^{(l)} \right) W_{\text{down}}^{(l)},
    \label{eq:ffn}
\end{equation}
where \( W_{\text{up}}^{(l)},\; W_{\text{gate}}^{(l)} \in \mathbb{R}^{d \times d_{\text{ff}}/2} \), and \( W_{\text{down}}^{(l)} \in \mathbb{R}^{d_{\text{ff}}/2 \times d} \). Here, \textit{SwiGLU} activation function is defined as:

\begin{equation}
    \text{SwiGLU}(a, b) = \text{SiLU}(a) \odot b,
    \label{eq:swiglu}
\end{equation}
where \( a = x W_{\text{up}}^{(l)} \) and \( b = x W_{\text{gate}}^{(l)} \), both of shape \( \mathbb{R}^{n \times d_{\text{ff}}/2} \). This two-branch formulation matches the implementation in \textit{DeepSeek-R1}, where the gating branch \( W_{\text{gate}}^{(l)} \) is activated and multiplied elementwise with the value branch \( W_{\text{up}}^{(l)} \). The \textit{SiLU} activation~\citep{elfwing2017silu} is applied elementwise as \( \text{SiLU}(x) = x / (1 + e^{-x}) \), and \( \odot \) denotes elementwise multiplication.

Finally, the \textit{residual connection} yields the output of the feedforward sub-layer:

\begin{equation}
    \boldsymbol{H}^{(l)}_{\text{out}} = \boldsymbol{H}^{(l)}_{\text{mid}} + \text{FFN}^{(l)}\left( \text{RMSNorm}(\boldsymbol{H}^{(l)}_{\text{mid}}) \right).
    \label{eq:ffn_output}
\end{equation}

\subsubsection{Output Layer}

At each decoding step $t$, the model uses the hidden state vector at the final position of the last Transformer block, denoted as $\mathbf{h}_t \in \mathbb{R}^d$. This vector summarizes the context up to step $t$ and serves as the input for next-token prediction.

To predict the next token $y_{t}$, the model applies a linear projection followed by a softmax over the vocabulary:

\begin{equation}
\hat{\boldsymbol{y}}_{t} = \text{softmax} \left( \frac{\mathbf{h}_t W_{\text{out}}^\top + \boldsymbol{b}_{\text{out}}}{\beta} \right),
\label{eq:output_layer}
\end{equation}
where $W_{\text{out}} \in \mathbb{R}^{|\mathcal{V}| \times d}$ is the output projection matrix, $\boldsymbol{b}_{\text{out}} \in \mathbb{R}^{|\mathcal{V}|}$ is a learned bias vector, and $|\mathcal{V}|$ is the vocabulary size. The scalar $\beta > 0$ is the temperature parameter that controls the sharpness of the output distribution: lower $\beta$ produces sharper (more peaked) probabilities, while higher $\beta$ increases randomness. In short, a higher temperature encourages the model to be more ``creative'' by increasing the randomness in generating the next token.

The resulting vector $\hat{\boldsymbol{y}}_{t} \in \mathbb{R}^{|\mathcal{V}|}$ defines a categorical probability distribution over the vocabulary for the next token.

\subsubsection{Inference}

Once the model computes the probability distribution $\hat{\boldsymbol{y}}_{t}$ over the vocabulary (as defined in Equation~\ref{eq:output_layer}), a decoding strategy is used to select the next token \( y_{t} \). Several decoding strategies are commonly used:

\begin{itemize}
    \item \textbf{Greedy decoding}: Select the token with the highest probability.
    \item \textbf{Sampling}: Draw a token randomly from the full distribution defined by $\hat{\boldsymbol{y}}_{t}$.
    \item \textbf{Top-$k$ filtering}: Truncate the distribution to the $k$ most probable tokens, then sample from the resulting subset.
    \item \textbf{Top-$p$ filtering} (nucleus filtering): Select the smallest set of tokens whose cumulative probability exceeds a threshold $p$, and sample from this set.
\end{itemize}

By default, \textit{DeepSeek-R1} uses sampling with a temperature of 1.0 and top-$k$ filtering with $k = 50$, generating each token by sampling from the truncated and renormalized distribution.

\subsubsection{LoRA Fine-Tuning}

\paragraph{LoRA Modeling}

\begin{figure}[htbp]
    \centering
    \includegraphics[width=0.7\textwidth]{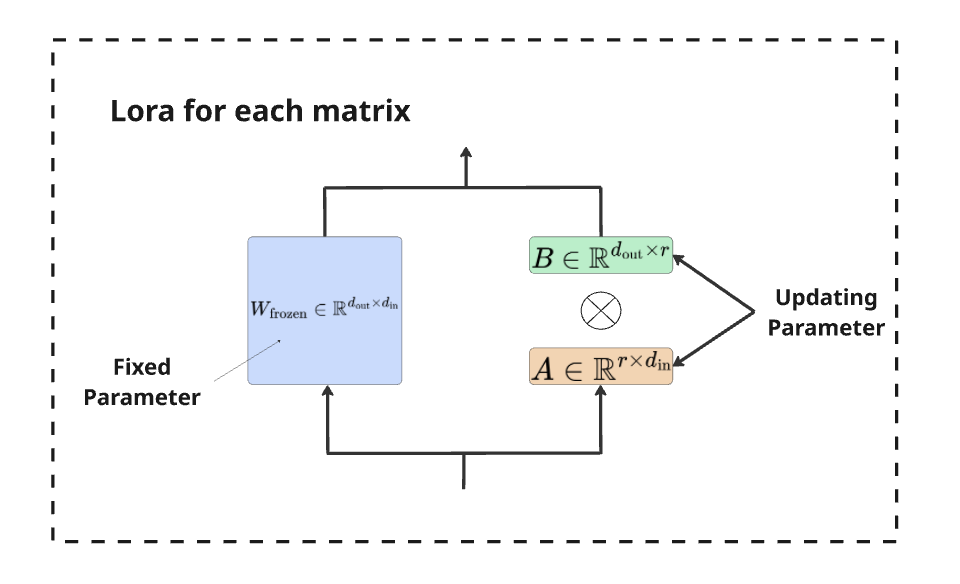}
    \caption{
        LoRA adaptation applies to a single projection matrix. 
        The original weight matrix $W_\mathrm{frozen} \in \mathbb{R}^{d_\mathrm{out} \times d_\mathrm{in}}$ remains unchanged, 
        while trainable matrices $A \in \mathbb{R}^{r \times d_\mathrm{in}}$ and $B \in \mathbb{R}^{d_\mathrm{out} \times r}$ 
        introduce a low-rank update via $\Delta W = B A$. 
        The effective weight is $W_\mathrm{frozen} + \Delta W$.
    }
    \label{fig:lora_matrix}
\end{figure}

We apply \textit{Low-Rank Adaptation} (\textit{LoRA}) ~\citep{hu2022lora} to selected linear transformations within the Transformer decoder. Let \( W_\text{frozen} \in \mathbb{R}^{d_{\text{out}} \times d_{\text{in}}} \) be a frozen weight matrix (e.g., a projection matrix in self-attention or feedforward layers), which contains pre-trained knowledge for the foundation model. \textit{LoRA} introduces a low-rank trainable perturbation \( \Delta W \in \mathbb{R}^{d_{\text{out}} \times d_{\text{in}}} \) defined as:
\begin{equation}
\Delta W = BA, \quad A \in \mathbb{R}^{r \times d_{\text{in}}}, \quad B \in \mathbb{R}^{d_{\text{out}} \times r}, \quad r \ll \min(d_{\text{in}}, d_{\text{out}})
\label{eq:lora_delta}
\end{equation}

Optionally, a scaling factor \( \alpha \in \mathbb{R} \) is applied:
\begin{equation}
W = W_\text{frozen} + \frac{\alpha}{r} \Delta W = W_\text{frozen} + \frac{\alpha}{r} BA
\label{eq:lora_scaled}
\end{equation}

This modified transformation \( W \) is used during decoding, while \( W_\text{frozen} \) remains unchanged. Only the low-rank matrices \( A, B \) are updated to modify parameters based on domain training data.

Let \( \mathcal{D} = \{ (\boldsymbol{x}^{(i)}, \boldsymbol{y}^{(i)}) \}_{i=1}^{N} \) be the training dataset. The \textit{LoRA}-parameterized model defines the same conditional distribution:
\begin{equation}
P(\boldsymbol{y} \mid \boldsymbol{x}; \boldsymbol{\theta}_{\text{LoRA}}) = \prod_{t=1}^{T} P(y_t \mid \boldsymbol{y}_{<t}, \boldsymbol{x}; \boldsymbol{\theta}_{\text{LoRA}})
\label{eq:lora_autoreg}
\end{equation}
with loss function:
\begin{equation}
\mathcal{L}_{\text{LoRA}}(\boldsymbol{\theta}_{\text{LoRA}}) = - \sum_{i=1}^{N} \sum_{t=1}^{T^{(i)}} \log P\left(y_t^{(i)} \mid \boldsymbol{y}^{(i)}_{<t}, \boldsymbol{x}^{(i)}; \boldsymbol{\theta}_{\text{LoRA}}\right)
\label{eq:lora_loss}
\end{equation}
where \( \boldsymbol{\theta}_{\text{LoRA}} \) is the collection of all inserted low-rank parameters.

\begin{figure}[htbp]
    \centering
    \includegraphics[width=0.7\textwidth]{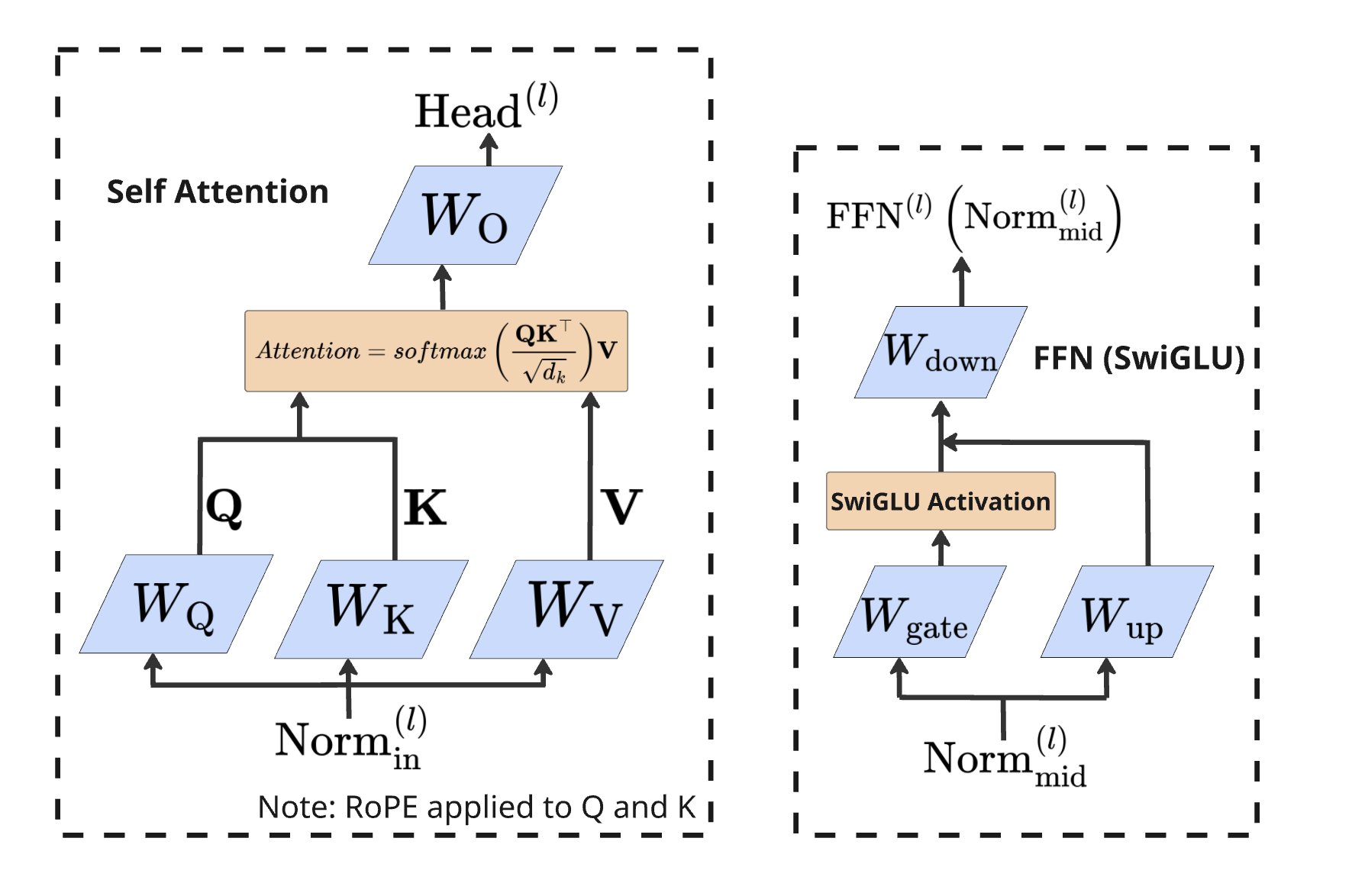}
    \caption{
        Self-attention module and FFN in transformer. 
        Each matrix represented by a light blue parallelogram corresponds to a learnable projection matrix within the transformer block. These matrices are potential targets for LoRA adaptation, where low-rank updates can be applied individually to each matrix.
    }
    \label{fig:transformer_lora_targets}
\end{figure}

\paragraph{LoRA Placement Options}

We apply \textit{LoRA} to selected projection matrices in each Transformer block, following the architecture described in Figure \ref{fig:transformer_lora_targets}. Each block contains a multi-head causal self-attention module and a FFN. Candidate insertion points include:

\paragraph{1. Attention Projections.}
For each attention head \( i = 1, \ldots, m \), we may apply LoRA to the query, key, and value projections:

\begin{equation}
W_{i, \text{Q}} = W_{i, \text{Q}, \text{frozen}} + \Delta W_{i, \text{Q}}, \quad \Delta W_{i, \text{Q}} = B_{i, \text{Q}} A_{i, \text{Q}}
\label{eq:lora_query}
\end{equation}

\begin{equation}
W_{i, \text{K}} = W_{i, \text{K}, \text{frozen}} + \Delta W_{i, \text{K}}, \quad \Delta W_{i, \text{K}} = B_{i, \text{K}} A_{i, \text{K}}
\label{eq:lora_key}
\end{equation}

\begin{equation}
W_{i, \text{V}} = W_{i, \text{V}, \text{frozen}} + \Delta W_{i, \text{V}}, \quad \Delta W_{i, \text{V}} = B_{i, \text{V}} A_{i, \text{V}}
\label{eq:lora_value}
\end{equation}

\paragraph{2. Attention Output Projection.}
After the attention heads are concatenated, an output projection is applied:
\begin{equation}
W_{\text{O}} = W_{\text{O}, \text{frozen}} + \Delta W_{\text{O}}, \quad \Delta W_{\text{O}} = B_{\text{O}} A_{\text{O}}
\label{eq:lora_output}
\end{equation}

\paragraph{3. Feedforward Projections.}
Each FFN consists of two linear layers. LoRA may be applied to one or both:

\begin{equation}
W_{\text{up}} = W_{\text{up}, \text{frozen}} + \Delta W_{\text{up}}, \quad \Delta W_{\text{up}} = B_{\text{up}} A_{\text{up}}
\label{eq:lora_ffn1}
\end{equation}

\begin{equation}
W_{\text{gate}} = W_{\text{gate}, \text{frozen}} + \Delta W_{\text{gate}}, \quad \Delta W_{\text{gate}} = B_{\text{gate}} A_{\text{gate}}
\label{eq:lora_ffn_gate}
\end{equation}

\begin{equation}
W_{\text{down}} = W_{\text{down}, \text{frozen}} + \Delta W_{\text{down}}, \quad \Delta W_{\text{down}} = B_{\text{down}} A_{\text{down}}
\label{eq:lora_ffn2}
\end{equation}

\paragraph{Default Configuration.}
By default, \textit{LoRA} is applied to the query and value projections (\( W^Q \), \( W^V \)) in each Transformer block, as these components most directly influence the attention mechanism. Other insertion points can be optionally enabled depending on task complexity and computation budget.

\section{Empirical Implementation} \label{sec:EI}

\subsection{Vehicle Warranty Dataset}  \label{sec:data}

We conduct our study on a large-scale proprietary cleaned corpus of approximately 2.0 million real-world automotive warranty claims provided by PCMI Corporation\footnote{https://www.pcmicorp.com/} which is a global leader in providing software solutions for insurance companies, auto manufacturers, and Third-Party Administrators. We are able to acquire the databases that contain the entire lifecycle of a warranty, from the moment a customer signs a contract at a dealership to the moment a repair shop files a claim. Each claim corresponds to a repair event but may span multiple claim events linked by several identifiers (ClaimID, ContractID, VIN prefix) across dozens of databases, reflecting complicated warranty administration workflows in which a single incident generates multiple diagnostic or repair records.

Each claim contains three free-text narratives written by technicians under operational time constraints: \textit{Complaint} (customer-reported issues and complaints), \textit{Cause} (mechanical diagnostic reasoning), and \textit{Correction} (the repair action performed). These narratives constitute the supervision signal used for model training and evaluation. There are also some structured metadata---such as make, model, year, and contract-level attributes---that support cross-coverage consolidation, repeated-visit detection, and stratified analysis across vehicle classes.

The raw corpus is characterized by substantial real-world noise, a common artifact of industrial service logs. These irregularities, including inconsistent formatting and malformed entries, reflect the inherent complexity of data captured in live production environments, necessitating robust preprocessing strategies. For example, text fields contain dense domain-specific shorthand (e.g., ``r\&r'', ``no crank'', ``diag elec''), heterogeneous misspellings, incomplete or fragmentary sentences, and technician-specific lexical styles. Multi-coverage claims frequently include duplicated or partially copied narratives, and administrative comments (e.g., billing or routing notes) often appear intermixed with technical content. To ensure semantic consistency, we develop a deterministic preprocessing pipeline to ensure consistency, which includes five stages:
\begin{enumerate}
    \item Removal of duplicated or mechanically copied textual fragments;
    \item Construction of a domain-specific rewrite dictionary to normalize abbreviations and shorthand;
    \item Correction of high-frequency misspellings and lexical variants based on corpus statistics;
    \item Elimination of administrative text unrelated to diagnostic or repair reasoning;
    \item Filtering of incomplete or severely degraded narratives that fail to provide a meaningful supervisory signal.
\end{enumerate}

After preprocessing, each claim is represented as a consolidated Complaint--Cause--Correction tuple paired with vehicle metadata. We partition the dataset into training, validation, and held-out test sets in accordance with the data-sharing agreement.

Statistically, the dataset exhibits heavy-tailed repair frequencies, a long-tailed technician-driven vocabulary, and significant heterogeneity across manufacturers and vehicle classes. These characteristics create a challenging evaluation environment for assessing whether domain-aligned fine-tuning produces format-stable, machine-extractable outputs, and maintains consistent outputs that adhere to established business logic.

\subsection{Task Setup and Masked Autoregressive Training}
\label{subsec:task-setup-masked-training}

To implement the conditional objective defined in Section~\ref{sec:formulation} under a decoder-only autoregressive architecture, each training observation is represented as a single token sequence consisting of an instruction segment describing the claim context and a response segment containing the corresponding technician-recorded corrective action.

After formatting and tokenization, each observation is represented as a single autoregressive sequence
\begin{equation}
\boldsymbol{z} = (z_1, z_2, \ldots, z_L) = \text{Concat}(\boldsymbol{x}, \boldsymbol{y}),
\end{equation}
where $\boldsymbol{x} = (x_1, \ldots, x_n)$ denotes instruction tokens (Complaint--Cause), $\boldsymbol{y} = (y_1, \ldots, y_T)$ denotes corrective-action tokens (Correction), and $L = n + T$ is the total sequence length.

To restrict learning exclusively to corrective-action generation, we define a binary supervision mask $\boldsymbol{m} \in \{0,1\}^L$ such that
\begin{equation}
m_t =
\begin{cases}
0, & 1 \leq t \leq n, \\
1, & n < t \leq n + T,
\end{cases}
\end{equation}
and compute training loss only on response tokens.

Under this design, the masked training objective is defined as
\begin{equation}
\mathcal{L}_{\text{mask}}(\boldsymbol{\theta}_{\text{LoRA}})
= - \sum_{i=1}^{N} \sum_{t=1}^{L^{(i)}} m_t^{(i)} 
\log P \big(z_t^{(i)} \mid \boldsymbol{z}_{<t}^{(i)}; \boldsymbol{\theta}_{\text{LoRA}} \big),
\end{equation}
where $P(\cdot \mid \cdot; \boldsymbol{\theta}_{\text{LoRA}})$ denotes the autoregressive distribution induced by the LoRA-parameterized model, and $L^{(i)} = n^{(i)} + T^{(i)}$ is the length of the $i$-th training observation.

All sequences are truncated or padded to a maximum length of 2048 tokens. Instruction and padding tokens are excluded from the loss through label masking. This supervision scheme ensures that fine-tuning directly reshapes the conditional output distribution over corrective actions, rather than modeling prompt syntax or formatting artifacts.

\subsection{LoRA-Based Model Adaptation}

We perform domain adaptation of the foundation model, DeepSeek-R1, using LoRA. All pretrained backbone parameters remain frozen to preserve pretrained knowledge, and only low-rank adapters are optimized. For each frozen projection matrix $W_{\text{frozen}}$, LoRA introduces a trainable low-rank perturbation
\begin{equation}
W = W_{\text{frozen}} + \frac{\alpha}{r} BA,
\end{equation}
where $A \in \mathbb{R}^{r \times d_{\text{in}}}$, $B \in \mathbb{R}^{d_{\text{out}} \times r}$, and $r \ll \min(d_{\text{in}}, d_{\text{out}})$. Only the collection of low-rank parameters $\boldsymbol{\theta}_{\text{LoRA}} = \{A, B\}$ is optimized under $\mathcal{L}_{\text{mask}}(\boldsymbol{\theta}_{\text{LoRA}})$.

We inject LoRA adapters into both attention and feedforward projection layers of each Transformer block, including the query, key, value, output, gate, up, and down projections, following the formalization in~\ref{sec:formulation}. This placement enables adaptation of both contextual alignment mechanisms and nonlinear transformation subspaces.

Training is performed by minimizing the masked negative log-likelihood objective
\begin{equation}
\boldsymbol{\theta}^*_{\text{LoRA}} = \arg \min_{\boldsymbol{\theta}_{\text{LoRA}}} \mathcal{L}_{\text{mask}}(\boldsymbol{\theta}_{\text{LoRA}}),
\end{equation}
using the AdamW optimizer under mixed-precision (FP16) training. The per-device batch size is 8 with gradient accumulation in 4 steps. All experiments are conducted for a single training epoch, with a learning rate of $6 \times 10^{-5}$.

We adopt a LoRA rank of $r = 32$ with scaling factor $\alpha = 32$ and zero dropout. More details for the implementation can be found in Appendix~\ref{appendix:implementation}. This configuration enables the model to acquire domain-specific operational distributions while maintaining a small trainable parameter footprint, supporting stable adaptation of a large foundation model under industrial data and governance constraints.

\subsection{Model Evaluation}
\label{sec:eval}

The evaluation of LLMs outputs is a non-trivial task that necessitates a departure from simple, deterministic validation metrics. Unlike traditional classification or regression tasks, LLM performance must be analyzed through a multidimensional lens to capture the nuances of natural language generation. A primary challenge lies in the discrepancy between lexical overlap and semantic validity. For instance, a model-generated response may lack exact string alignment with the ground-truth reference, yet remain semantically identical or factually correct. As the field shifts toward more sophisticated assessment strategies, practitioners must balance the trade-off between the scalability of automated evaluation metrics and the nuanced accuracy of human or model-based qualitative reviews.

\begin{table}[t]
\small
\centering
\caption{Taxonomy of  automated evaluation metrics used in this study}
\label{tab:metrics}
\begin{tabular}{p{4.2cm} p{4.8cm} p{7.1 cm}}
\toprule
\textbf{Metric} & \textbf{Family} & \textbf{What it measures} \\
\midrule
Edit distance similarity & Surface-level similarity & Character- or token-level similarity derived from edit distance between prediction and reference \\
Unigram precision & Surface-level similarity & 1-gram overlap precision between prediction and reference \\
Bigram precision & Surface-level similarity & 2-gram overlap precision between prediction and reference \\
Sentence BLEU & Surface-level similarity & Sentence-level n-gram overlap between prediction and reference (BLEU) \\
\midrule
TF-IDF cosine similarity & Lexical--semantic similarity & Weighted lexical overlap in a bag-of-words vector space \\
BERT cosine similarity & Lexical--semantic similarity & Contextual sentence-level representation similarity \\
BERTScore (F1) & Lexical--semantic similarity & Semantic correspondence via contextualized token embeddings (F1 aggregation) \\
\midrule
BLEURT (normalized) & Model-based evaluator & Learned neural evaluator score (normalized) trained to approximate human judgment \\
LLM-as-a-Judge score & Model-based evaluator & LLM-based semantic evaluation under a structured scoring rubric aligned with human judgments \\
\bottomrule
\end{tabular}
\end{table}

Implementation details and formal definitions of all evaluation metrics discussed in Table \ref{tab:metrics} are provided in Appendix~\ref{appendix:Metrics}.

\subsubsection{Automated Evaluation Metrics}
\label{sec:automated_eval}

Automated evaluation is intended to approximate human judgment of semantic correctness and practical validity while enabling scalable and reproducible benchmarking. However, different automated metrics capture fundamentally different notions of similarity. To reflect this, we organize all evaluated metrics in Table \ref{tab:metrics} into three conceptual families: \emph{surface-level similarity}, \emph{lexical--semantic similarity}, and \emph{model-based evaluator}. This categorization allows us to analyze not only absolute metric values, but also how different classes of evaluators align with human assessments.

\paragraph{Surface-level similarity metrics.}
Surface-level metrics quantify explicit string overlap and local textual similarity between predicted and ground-truth references. These methods are sensitive to token identity, word order, and edit operations, and primarily measure whether a prediction reproduces the surface form of the reference.

We consider Edit distance similarity \citep{levenshtein1966binary}, n-gram precision \citep{papineni2002bleu}, and Sentence BLEU \citep{papineni2002bleu} as representatives of this family. Sentence BLEU and n-gram precision capture local lexical overlap at varying granularities, while Edit distance similarity measures the minimum number of insertions, deletions, and substitutions required to transform a prediction into its reference. Collectively, these metrics serve as strict baselines that emphasize form-level correspondence, but are known to be brittle under paraphrasing and terminology variation.

\paragraph{Lexical--semantic similarity metrics.}
Lexical--semantic metrics move beyond exact surface matching by embedding text into continuous vector spaces, where similarity reflects soft lexical and distributional semantic alignment. These methods are more tolerant to paraphrasing and synonym substitution, while still operating primarily at the representation-comparison level and depending on predefined embeddings.

We consider TF--IDF cosine similarity \citep{manning2008introduction}, BERT cosine similarity \citep{reimers2019sentence} and BERTScore \citep{zhang2020bertscore} in this category. TF--IDF cosine similarity captures weighted lexical overlap under a bag-of-words representation, whereas BERT cosine similarity compares contextualized sentence embeddings derived from pretrained transformers. BERTScore further extends this paradigm by computing token-level semantic alignment through maximum-matching in a contextual embedding space, aggregating precision, recall, and F1 scores. While these metrics partially account for semantic proximity, they remain fundamentally similarity-based and are agnostic to task-specific correctness or, in our case, the validity of corrective actions.

\paragraph{Model-based evaluators.}
Model-based evaluators do not rely on fixed or closed-form similarity functions. Instead, they employ neural models as black-box scoring functions that directly map a candidate--reference pair to a scalar score, with the objective of approximating human judgments of semantic alignment or quality. As LLMs continue to advance, such model-based evaluators have demonstrated increasingly strong empirical performance and have become widely adopted in recent evaluation frameworks.

We consider BLEURT \citep{sellam2020bleurt} and an LLM-as-a-Judge \citep{zheng2023judging} score as representative model-based evaluators. BLEURT is a task-specific evaluation model explicitly trained on synthetic perturbations and human ratings to predict human-aligned quality scores. In contrast, our LLM-as-a-Judge evaluator is based on \texttt{ChatGPT-4o-mini}, a general-purpose language model that is not specifically trained for evaluation but is prompted to act as a judge and score model outputs according to the same rubric used in human evaluation.

\begin{table}[t]
\centering
\caption{Human evaluation scoring guidelines used in all annotation tasks.}
\label{tab:human-guideline}
\begin{tabular}{p{1.2cm} p{2.5cm} p{11.5cm}}
\toprule
\textbf{Score} & \textbf{Meaning} & \textbf{Description} \\
\midrule
1.0 & Perfect Match &
The predicted actions fully and precisely cover the ground-truth actions.
No missing or incorrect actions. \\

0.8 & Strong Match &
The predicted actions mostly cover the ground truth, with only minor omissions
or paraphrasing and no major errors. \\

0.6 & Partial Match &
The prediction covers some key actions but misses important parts.
Still related and somewhat helpful. \\

0.4 & Weak Match &
Very limited overlap. Only a few actions (or fragments) are relevant
to the ground truth. \\

0.2 & Minimal Match &
Prediction is largely irrelevant or incorrect, with almost no valid overlap
with the ground truth. \\

0.0 & No Match &
Completely unrelated or invalid prediction, with no correspondence to the
ground truth. \\
\bottomrule
\end{tabular}
\end{table}

\subsubsection{Human Evaluation}
\label{sec:human_eval}

Human evaluation is considered the gold standard, especially for experts in the experience domain, because automated metrics provide no intrinsic guarantee of semantic correctness. Different automated evaluators capture fundamentally different notions of similarity, and in the corrective action setting, surface-level overlap can diverge substantially from whether a predicted action is semantically correct. As a result, purely automatic scoring cannot determine whether a model output is truly valid or merely lexically similar. We therefore use human judgment as a semantic reference for interpreting model performance and for contextualizing the behavior of automated evaluation metrics.

Human annotators score each prediction based on its \emph{semantic alignment} with the reference corrective action. The goal is to assess whether the predicted action conveys the same underlying meaning as the reference, rather than whether it matches the reference wording. In particular, annotators consider whether the core corrective intent is preserved and whether the predicted action is relevant to the described claim context.

Since semantic judgment is subjective and sensitive to implicit decision boundaries, we standardize human evaluation using a unified discrete ordinal rubric (Table~\ref{tab:human-guideline}). The rubric defines six quality levels that capture increasing degrees of semantic alignment, ranging from a complete mismatch to fully aligned actions. Each level is associated with explicit criteria, providing a shared evaluation scale across observations. This structured rubric reduces ambiguity in annotation and enables systematic comparison between human scores and automated evaluation metrics.

\subsubsection{Aligning Automated Metrics with Human Judgment}
\label{sec:correlation_analysis}

\begin{table}[t]
\centering
\caption{Correlation between different automated evaluation metrics and human judgments.}
\label{tab:metric_correlation}
\begin{tabular}{lccc}
\hline
\textbf{Metric} & \textbf{Chatterjee $\xi$} & \textbf{Spearman $\rho$} & \textbf{Kendall $\tau$} \\
\hline
BERT cosine similarity & \textbf{0.692} & \textbf{0.733} & 0.607 \\
LLM-as-a-Judge score   & 0.606 & 0.724 & \textbf{0.678} \\
BLEURT (normalized)   & 0.576 & 0.703 & 0.566 \\
TF-IDF cosine similarity & 0.577 & 0.683 & 0.564 \\
BERTScore (F1)        & 0.581 & 0.676 & 0.537 \\
Edit distance similarity       & 0.530 & 0.655 & 0.526 \\
Unigram precision     & 0.400 & 0.579 & 0.495 \\
Sentence BLEU         & 0.495 & 0.579 & 0.467 \\
Bigram precision      & 0.426 & 0.530 & 0.438 \\
\hline
\end{tabular}
\end{table}

While automated evaluation metrics are fast, consistent, and easy to scale, they do not inherently reflect how humans understand language and domain expertise. In contrast, human judgment directly captures semantic correctness and meaning-level validity, making it the most faithful reference for evaluating predicted actions. However, human evaluation is costly, time-consuming, and difficult to apply at scale. As a result, automated evaluation metrics and human evaluation should not be viewed as competing alternatives, but as complementary components of an evaluation framework: human judgment defines the semantic reference, while automated evaluation metrics provide scalable approximations.

From this perspective, automated evaluation metrics are meaningful only to the extent that they reflect human judgment. We therefore treat human evaluation as a semantic gold standard and assess automated evaluation metrics by examining how closely their induced rankings of predicted actions align with those given by human evaluators. Correlation analysis serves as the measurement for this purpose, enabling us to quantify the degree to which different automated evaluation metrics preserve human preference structure.

Because we do not assume a linear relationship between metric scores and human judgments, we adopt a multi-perspective correlation framework. Specifically, we compute Chatterjee’s rank correlation coefficient $\xi$ \citep{chatterjee2021new}, Spearman's $\rho$ \citep{spearman1904proof}, and Kendall’s $\tau$ \citep{kendall1938new} between automated evaluation metrics-induced rankings and human rankings. These statistics respectively capture general dependence, monotonic association, and pairwise ranking consistency, providing a comprehensive view of how faithfully each automated evaluation metric approximates human evaluation.

Across all automated evaluation metrics, BERT cosine similarity and the LLM-as-a-Judge method consistently demonstrate the strongest and most stable alignment with human judgments, as indicated by correlation measures. This indicates that evaluation paradigms grounded in contextual semantic representations and learned semantic judgments more closely reflect human notions of correctness than surface-level similarity metrics. Accordingly, we adopt these two metrics as our primary automated evaluation suite for subsequent experiments, treating them as the most reliable, scalable proxies for human evaluation.


\section{Model Governance} \label{sec:results}

\subsection{Overview}
\label{sec:Experiment Overview}

To examine how domain-specific fine-tuning alters the behavior of general-purpose LLMs on the structured warranty claim corrective action prediction task, we construct a controlled set of four commonly used LLM-based model variants summarized in Table~\ref{tab:model_config}. These variants are organized to contrast general-purpose models with a domain-adapted counterpart, while carefully controlling for prompting strategy and base foundation model family. This design allows us to isolate the effect of domain adaptation and examine how it interacts with structured prompting and model choice.

\begin{table}[!htbp]
\centering
\small
\caption{Model configurations.}
\label{tab:model_config}
\begin{tabular}{c p{2.5cm} c c c p{7cm}}
\toprule
Model & Base model & Scale & Prompt & Fine-tuned & Training data \\
\midrule
M1 & DeepSeek-R1     & 8B &  $\times$  & $\times$   & General-purpose data\\
M2 & DeepSeek-R1     & 8B & $\checkmark$ & $\times$   & General-purpose data\\
M3 & Qwen-Instruct   & 7B & $\checkmark$ & $\times$   & General-purpose data\\
M4 & DeepSeek-R1     & 8B & $\times$   & $\checkmark$ & General-purpose + Domain-specific data\\
\bottomrule
\end{tabular}
\end{table}

M1 serves as a general-purpose baseline without explicit format constraints or domain adaptation. M2 introduces explicit format prompting while holding the base model fixed, enabling us to examine how structured prompting alone influences usability and accuracy. M3 replaces the base model with an instruction-tuned alternative under the same prompting strategy, allowing us to assess the impact of foundation model family and alignment style. M4, trained following the procedure described in Section~\ref{sec:EI}, further incorporates domain-specific fine-tuning on the warranty corpus, enabling a direct comparison between general-purpose and domain-adapted models. All model variants are evaluated under identical experimental conditions. The same test split, input cases, and decoding parameters are used throughout all experiments. Except for the explicit format constraint imposed in M2 and M3, the prompt structure and inference pipeline remain fixed across models. This controlled experimental design ensures that observed performance differences can be attributed to intrinsic model behavior rather than evaluation artifacts.

We evaluate model behavior along two complementary dimensions: \textit{usability} and \textit{accuracy}. Usability captures whether model outputs can be reliably consumed by downstream systems and therefore support automated pipelines. It encompasses both format compliance and response validity. Format compliance measures whether outputs strictly follow the predefined schema and are therefore machine-parsable. Response validity assesses whether parsable outputs contain identifiable and actionable information without exceeding the model generation limit, indicating that the response can remain informative and practically useful, even when its surface form is imperfect. Accuracy is evaluated separately and only on valid outputs. It measures how well extracted actions align with true references, capturing task-level semantic correctness. Together, this dual-dimensional evaluation framework allows us to analyze not only whether domain-specific fine-tuning improves end-task accuracy, but also whether it improves the practical usability of model outputs for structured automation. During error inspection, we find that some apparent prediction errors are driven by data quality issues and ambiguity in the warranty narratives, which cannot be fully eliminated despite extensive data preprocessing. To account for this factor, we additionally construct and evaluate on a manually verified high-quality subset, defined as \textit{HQ}, which is first selected using \texttt{ChatGPT-4o-mini} and then human-validated. It isolates cases with clear and semantically consistent corrective-action signals. We formally define our usability and accuracy as follows:

\paragraph{Definition} Given an evaluation set $\mathcal{D}=\{{x}^{(i)}\}_{i=1}^N$ with model outputs $\{y^{(i)}\}_{i=1}^N$ and a predefined output schema $\mathcal{S}$, 

Format compliance is defined as:
\[
\textbf{Format} = \frac{1}{N} \sum_{i=1}^N \mathbb{I}\!\left(y^{(i)} \text{ strictly conforms to } \mathcal{S}\right),
\]
Response validity is defined as:
\[
\textbf{Validity} = \frac{1}{N} \sum_{i=1}^N \mathbb{I}\!\left(y^{(i)} \text{ is parsable with actionable content}\right),
\]

Accuracy on valid outputs is defined as:
\[
\textbf{Acc(Valid)}
= \frac{1}{|\mathcal V|}
\sum_{i \in \mathcal V}
\mathbb{I}\!\left(
\mathrm{Sim}_{\text{Human}}(\hat{y}^{(i)}, y^{(i)}) \ge 0.8
\right),
\]
where $\mathcal V = \{ i \mid y^{(i)} \text{ is valid} \}$, $\hat{y}^{(i)}$ denotes the model-predicted corrective action, $y^{(i)}$ denotes the true reference, and $\mathrm{Sim}_{\text{Human}}(\cdot,\cdot)$ denotes the score assigned by human annotators following the scoring guideline.

Accuracy on the high-quality subset is defined as:
\[
\textbf{Acc(HQ)}
= \frac{1}{|\mathcal H|}
\sum_{i \in \mathcal H}
\mathbb{I}\!\left(
\mathrm{Sim}_{\text{Human}}(\hat{y}^{(i)}, y^{(i)}) \ge 0.8
\right),
\]
where $\mathcal H = \{ i \in \mathcal V \mid i \text{ is labeled as high-quality} \}$ denotes the manually verified high-quality subset.

Concrete examples illustrating format compliance and response validity are provided in Appendix~\ref{appendix:format}.

\begin{table*}[!htbp]
\centering
\caption{Main results on usability and accuracy.}
\label{tab:main_results}
\small
\setlength{\tabcolsep}{6pt}
\renewcommand{\arraystretch}{1.25}

\begin{tabular}{lcccc}
\toprule
& \multicolumn{2}{c}{\textbf{Usability}} 
& \multicolumn{2}{c}{\textbf{Accuracy}} \\
\cmidrule(lr){2-3} \cmidrule(lr){4-5}

\textbf{Model} 
& \textbf{Format (\%)} 
& \textbf{Validity (\%)} 
& \textbf{Acc (Valid) (\%)} 
& \textbf{Acc (HQ) (\%)} \\
\midrule

M1: DeepSeek-R1
& 0.0 & 8.0 & 56.0 & 64.3 \\

M2: DeepSeek-R1+ Prompt
& 6.5 & 29.5 & 64.4 & 71.2 \\

M3: Qwen-Instruct+ Prompt
& 86.5 & 97.5 & 62.1 & 68.4 \\

M4: DeepSeek-R1+ Fine-tune
& \textbf{100.0} & \textbf{100.0} & \textbf{81.5} & \textbf{92.0} \\

\bottomrule
\end{tabular}
\end{table*}


Table~\ref{tab:main_results} summarizes the performance of four LLM-based models evaluated across the dual dimensions of usability and accuracy. To ensure empirical rigor and the reliability of our findings, all results are benchmarked against a gold-standard dataset of $N=200$ human-verified warranty claim cases. The results demonstrate that the fine-tuned model achieves optimal performance in both format compliance and validity, while simultaneously yielding a substantial increase in accuracy. These findings suggest that fine-tuning facilitates more than mere surface-level adherence to output conformity; it effectively enables the base model to internalize task-specific domain knowledge latent within the training data.

\begin{figure*}[!t]
  \centering

  \begin{subfigure}{0.9\textwidth}
    \centering
    \includegraphics[width=\linewidth]{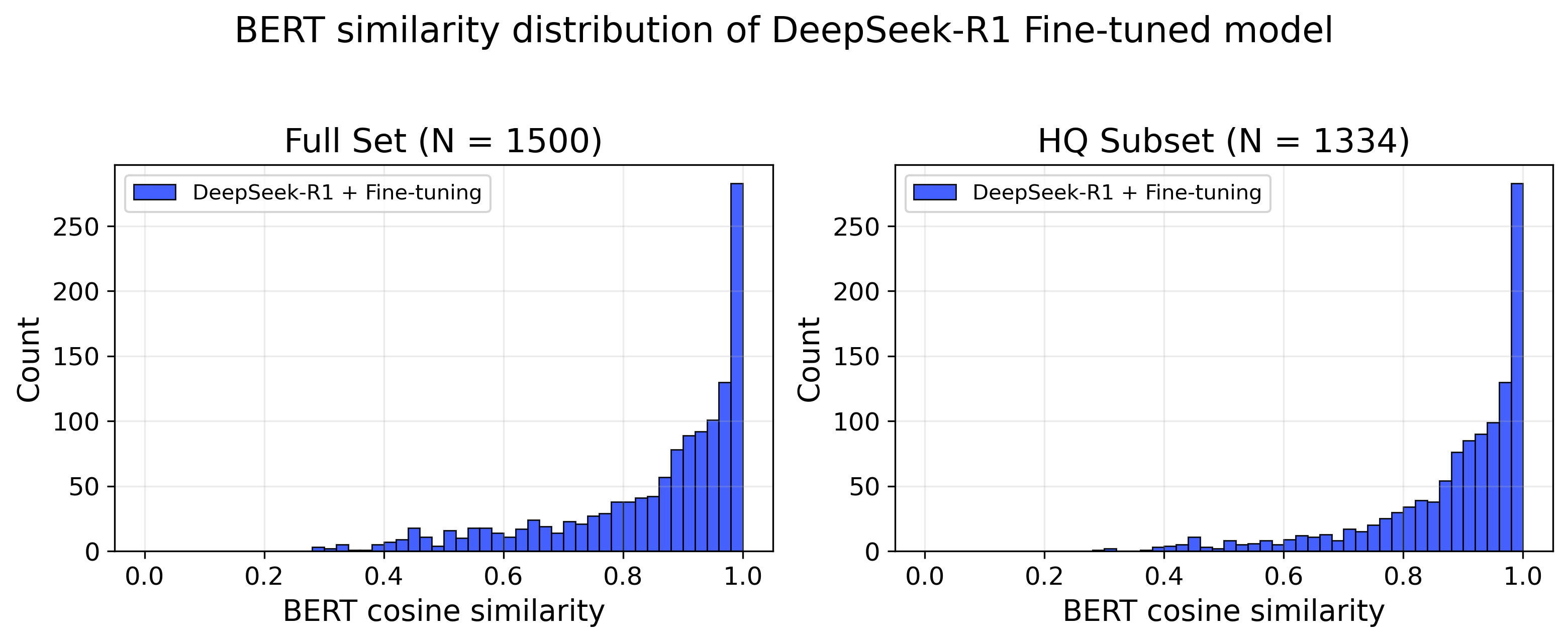}
    \caption{DeepSeek-R1 Fine-tuned model performance on the full evaluation set and the HQ subset.}
    \label{fig:bert_ft}
  \end{subfigure}

  \vspace{8pt}

  \begin{subfigure}{0.9\textwidth}
    \centering
    \includegraphics[width=\linewidth]{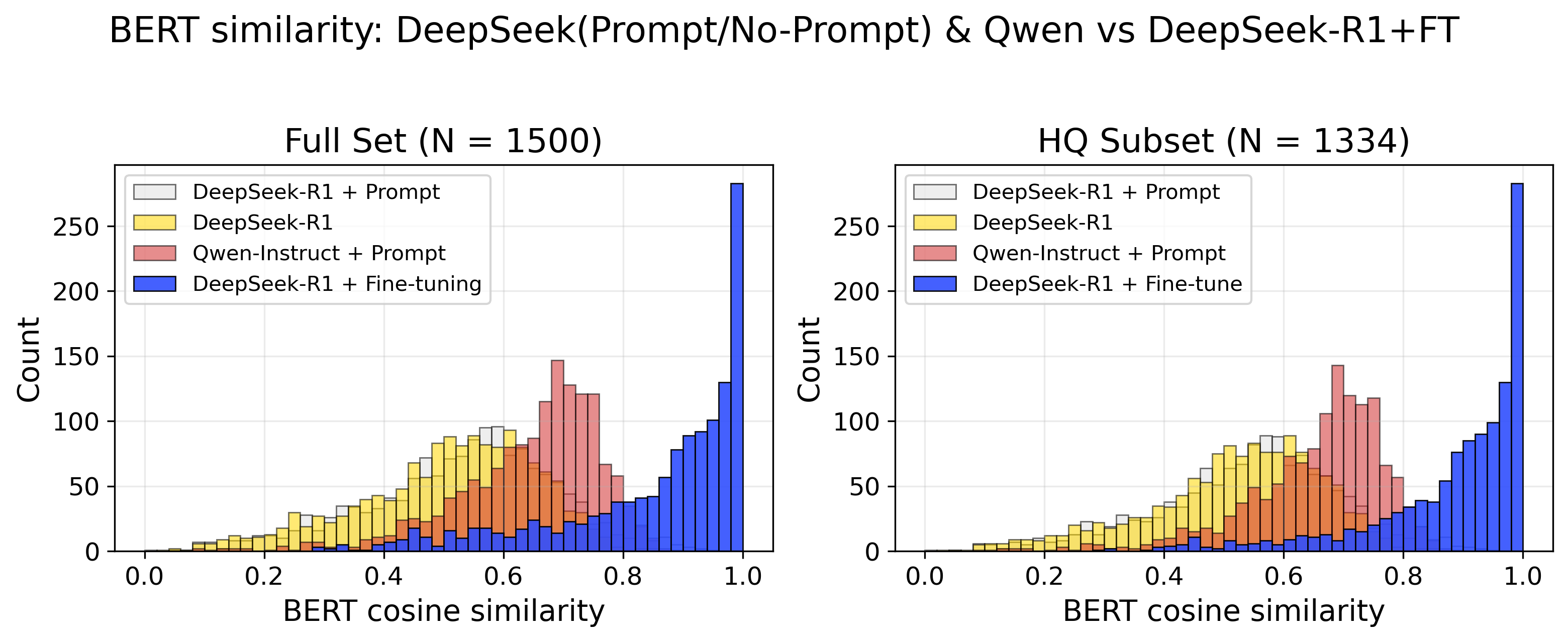}
    \caption{Comparison of M1-M4 on the full evaluation set and the HQ subset.}
    \label{fig:ft_vs_all}
  \end{subfigure}

  \caption{Semantic similarity distributions measured by BERT cosine similarity.}
  \label{fig:semantic_distributions}
\end{figure*}

Extending the analysis beyond aggregate usability and accuracy metrics, we examine the distributional changes in semantic similarity induced by domain-specific fine-tuning. As illustrated in Figure~\ref{fig:bert_ft}, the BERT cosine similarity distribution of the fine-tuned model exhibits reduced density in the lower similarity regime (0.4–0.6) when restricted to the HQ subset compared to the full evaluation set. Despite this minor difference, the overall distributional profiles remain largely congruent. This phenomenon is primarily attributed to the exclusion of ambiguous or low-quality corrective actions within the HQ subset, resulting in cleaner ground-truth references, rather than a structural divergence in model behavior. We further contrast the fine-tuned model against all the other models. As shown in Figure~\ref{fig:ft_vs_all}, the fine-tuned model exhibits a systematic rightward shift in the semantic similarity distribution compared to all three other models on both full evaluation set and HQ subset. This shift signifies that domain-specific fine-tuning effectively optimizes the base model to generate outputs with higher semantic alignment to expert human references, significantly outperforming general-purpose instruction-tuned or prompted baselines in this specialized task.

We subsequently investigate the mechanisms driving these observed behaviors in Section~\ref{sec:structural_controllability} and Section~\ref{sec:Accuaracy}.

\subsection{Usability: Format Compliance and Structural Controllability}
\label{sec:structural_controllability}

As indicated in Table~\ref{tab:main_results}, four LLM-based models exhibit sharply divergent usability, revealing a clear progression in format compliance as increasingly rigorous alignment strategies are applied. 

In the absence of task-specific alignment, DeepSeek-R1 operates in an unconstrained, general-purpose mode characterized by excessive verbosity and stylistic instability. Its outputs rarely adhere to the required schema, thereby obstructing downstream automation processes. Notably, human evaluation, which is necessitated by the near-total absence of machine-parsable formatting, reveals that only 8\% of its responses yield extractable, actionable information, with respect to the validity metric. These results underscore that general-purpose base models, when deployed without structural supervision, are fundamentally incompatible with automated actuarial workflows that demand deterministic, machine-parsable outputs. Furthermore, this provides critical empirical evidence explaining why the direct application of commercial, general-purpose APIs often leads to systemic failures and unstable performance in specialized industrial contexts.

The implementation of explicit, format-enforcing prompt engineering shifts the model into a weak, structurally controlled mode. The proportion of format-compliant outputs increases to 6.5\% and actionable outputs increases to 29.5\%, indicating that prompting can partially steer the model toward task relevance. However, structural controllability remains weak with simple prompt engineering. Despite stringent surface-level constraints, the model frequently reverts to free-form or reasoning-style generations that deviate from the required schema, reflecting the limited ability of prompt engineering to override generative patterns and the stochastic nature of a base model.

The instruct-tuned Qwen model demonstrates a marked shift in performance, characterized by a sharp increase in strict format compliance to 86.5\%, and a validity rate of 97.5\%. These results confirm that instruction tuning significantly enhances the model's ability to maintain structural adherence compared to baseline or prompt engineering. Nevertheless, occasional violations persist, indicating that instruction tuning effectively optimizes for probabilistic instruction-following behavior but does not enforce deterministic structural guarantees required for mission-critical, end-to-end claim automation. In an actuarial context, even a 13.5\% failure rate in formatting represents a significant bottleneck for downstream machine-parsable integration.

The fine-tuned DeepSeek-R1 model stands alone in consistently achieving full format compliance and structural controllability. Across the entire evaluated set, it achieves 100\% strict format compliance and 100\% validity. The complete absence of structural violations indicates that the fine-tuning process has effectively restricted the model's output to the predefined admissible schema.

Synthesizing these results, we establish a definitive hierarchy of structural control within the context of automated claim workflows: unconstrained generation in the base model, unstable partial control under prompt engineering, high-probability adherence but imperfect compliance under instruction tuning, and near-deterministic structural behavior under task-specific fine-tuning.

\subsection{Accuracy: Distribution Alignment}
\label{sec:Accuaracy}

Accuracy is evaluated exclusively on the valid set after filtering for format non-compliance. This methodological choice introduces a conservative bias favoring the non-fine-tuned models; while the fine-tuned model is evaluated on the full dataset, its counterparts are assessed only on the small fraction of the full dataset that successfully met format compliance. Despite vast discrepancies in usability, the three non-fine-tuned models exhibit surprisingly similar accuracy within their respective valid subsets, with scores clustering narrowly between 56.0\% and 64.4\%. This suggests that, despite substantial differences in base models or prompting strategies, non-fine-tuned models rely on a comparable, yet limited, semantic knowledge base inherited from pre-training. While pretrained foundation models acquire fragmented automotive repair knowledge from diverse internet sources that allow them to generate technically plausible responses, these outputs are derived from noisy and mixed distributions. Consequently, base and instruction-tuned models tend to encode generic plausibility priors rather than the specific operational conditional distributions that govern actual claim outcomes. 

In contrast, the fine-tuned model demonstrates a qualitatively different pattern in accuracy, increasing sharply to 81.5\% on the full set, clearly decoupling from all the non-fine-tuned models, which maintain performance levels only marginally better than a stochastic baseline. This separation becomes even more pronounced on the HQ subset. Under these denoised ground-truth conditions, the fine-tuned model reaches an accuracy of 92.0\%, whereas the remaining models continue to cluster between 64.3\% and 71.2\%. Such a leap represents a transition into a markedly higher predictive regime, confirming that fine-tuning enables the base model to internalize the rigorous logical structures required for downstream actuarial analysis. Notably, this advantage persists even when compared with SoTA commercial LLMs; despite their relatively large parameter scales, these general-purpose frontiers exhibit a persistent misalignment with specialized operational logic, as detailed in Appendix~\ref{appendix:Commercial}. This provides another compelling evidence that standard industry practices, such as relying solely on direct API calls to commercial LLMs, frequently prove inadequate across various complex scenarios.

More formally, following the formulation in Section~\ref{sec:formulation}, we characterize corrective-action prediction as a conditional sequence modeling task, as defined in Equation~\ref{eq:conditional}. Assume real-world claims operations generate outcomes according to an unknown data-generating process $P_{\text{data}}(\boldsymbol{y} \mid \boldsymbol{x})$. Pretraining a base model yields parameters $\boldsymbol{\theta}_{\text{pre}}$ by minimizing next-token prediction loss over massive, diverse internet source distribution $P_{\text{web}}(\boldsymbol{x}, \boldsymbol{y})$, producing a conditional probability $P(\boldsymbol{y} \mid \boldsymbol{x}; \boldsymbol{\theta}_{\text{pre}})$ that is well-calibrated for general-purpose. Prompting and instruction tuning modify the conditioning signal and, in the case of instruction tuning, adjust parameters on an instruction-following distribution $P_{\text{inst}}(\boldsymbol{x}, \boldsymbol{y})$. While these procedures improve stylistic compliance and local coherence, they do not directly optimize alignment with the operational process $P_{\text{data}}(\boldsymbol{y} \mid \boldsymbol{x})$. As a result, non-fine-tuned models frequently assign substantial probability mass to plausible but incorrect corrective action sequences that are common in internet narratives but invalid in actual claims operations.

Fine-tuning on domain-specific data makes the learning objective explicit. Given a domain dataset $\mathcal{D} = \{(\boldsymbol{x}^{(i)}, \boldsymbol{y}^{(i)})\}_{i=1}^{N}$ drawn from $P_{\text{data}}$, fine-tuning estimates parameters $\boldsymbol{\theta}_{\text{ft}}$ by minimizing the negative log-likelihood
\begin{equation}
\boldsymbol{\theta}_{\text{ft}} = \arg\min_{\boldsymbol{\theta}} \ \mathbb{E}_{(\boldsymbol{x},\boldsymbol{y})\sim p_{\text{data}}} \big[-\log P(\boldsymbol{y} \mid \boldsymbol{x}; \boldsymbol{\theta})\big],
\end{equation}
which is equivalent, up to an additive constant, to minimizing the conditional KL divergence
\begin{equation}
\boldsymbol{\theta}_{\text{ft}}
= \arg\min_{\boldsymbol{\theta}} \;
\mathbb{E}_{\boldsymbol{x} \sim P_{\text{data}}(\boldsymbol{x})}
\Big[ \mathrm{KL}\big(P_{\text{data}}(\boldsymbol{y} \mid \boldsymbol{x}) \,\|\, P(\boldsymbol{y} \mid \boldsymbol{x};\boldsymbol{\theta})\big) \Big].
\end{equation}

A full derivation of this equivalence is provided in Appendix~\ref{appendix:kl proof}. Thus, fine-tuning does not merely improve isolated predictions; it reallocates probability mass in $P(\boldsymbol{y} \mid \boldsymbol{x};\boldsymbol{\theta})$ so that empirical frequencies, hard constraints, and extreme asymmetries present in domain-specific data are reflected in the model's output distribution.

Specifically, this effect is most visible in rare-event cases. For any domain-specific event $A \subseteq \mathcal{Y}$ (e.g., ``predicting a repair-type action under a non-repairable condition''), fine-tuning shifts the induced event probability
\begin{equation}
P(A \mid \boldsymbol{x}; \boldsymbol{\theta}) = \sum_{\boldsymbol{y} \in A} P(\boldsymbol{y} \mid \boldsymbol{x}; \boldsymbol{\theta})
\end{equation}
toward its operational counterpart $P_{\text{data}}(A \mid \boldsymbol{x})$. Such distributional calibration is not fully captured by aggregated accuracy alone, but becomes evident when analyzing prediction populations and tail behavior. To illustrate this mechanism concretely, we present a domain-specific case study in Appendix~\ref{sec:tire_case_study}.

\subsection{Human-Centered Error Analysis and Failure Modes}
\label{sec:error_analysis}

To move beyond aggregate accuracy and examine model behavior at the granular observational level, we conducted a targeted error analysis on the held-out test set. Among the 1{,}500 evaluation cases, we identified 81 observations with LLM-as-a-Judge scores below 0.8 and subjected them to manual inspection. This human-centered analysis reveals that a substantial fraction of low-scoring cases do not correspond to actual prediction failures, but instead arise from surface-form variation or differences in output
granularity between model predictions and references.

For the remaining genuine errors, no single failure type dominates. Instead, errors are distributed across several categories, including omitted repairs, incorrect repair actions, and incorrect repair parts, without evidence of systematic collapse in any specific capability. Overall, the observed error landscape suggests stable model behaviors rather than structural deficiencies. These findings indicate that further performance gains are more likely to come from improved robustness and uncertainty-aware modeling, rather than targeting a single dominant failure mode.

Detailed error categories and representative examples are provided in Appendix~\ref{appendix:Error categoried}.

\section{Conclusion and Future Work} \label{sec:discussion}

This study develops a locally deployed and governance-aware claim-automation framework that integrates an open-source LLM into real-world automotive warranty claim workflows. Using a large-scale proprietary dataset comprising approximately 2.0 million historical claims, we fine-tune a DeepSeek-R1 8B foundation model via LoRA to generate structured corrective-action recommendations from unstructured Complaint--Cause narratives and other structured features. The proposed framework is designed to preserve data privacy, operate within an on-premises controlled environment, and support reproducibility and auditability, which are considered the key requirements in regulated insurance settings.

Empirically, our results demonstrate that domain-specific fine-tuning materially reshapes base model behavior in a way that prompt-only approaches do not reliably achieve. Fine-tuning improves structural controllability to ensure output format compliance, reshapes predictive distributions to reflect domain semantics on the targeted intermediate task, and yields near-exact matches to ground-truth corrective actions in a substantial fraction of evaluated cases. To assess both practical relevance and semantic correctness, we adopt a dual-dimensional evaluation approach that combines usability and accuracy, supported by automated semantic similarity metrics and human evaluation. This design enables a rigorous comparison between scalable automated metrics and human gold standard criteria. 

Importantly, while modern foundation models often exhibit impressive zero-shot or prompt-based competence on mechanical repair tasks, our experiments indicate that relying only on prompt engineering or general-purpose is insufficient for our claim automation task. Prompt engineering may influence how model outputs are expressed, but it does not alter which outputs receive probability mass. As a result, systems that rely solely on foundation models, regardless of architectural sophistication, remain fundamentally misaligned with real-world operational distributions.

More broadly, this study supports a pragmatic ``modular intermediate-task'' paradigm for the adoption of LLMs in actuarial and insurance applications. Rather than treating the LLM as an opaque end-to-end financial decision engine, we limit its role to a specific part of the pipeline where its linguistic understanding can help claim adjusters take an accelerated next-best course of action. In the future, we plan to design the downstream multi-step system to further automate claim management, for example, labor and parts estimation based on retrieval-augmented generation (RAG) applied on the internal database, reserving based on actuarial-informed neural networks \citep{cao2024assessing}. This modular design facilitates validation, interpretability, and principled integration with existing actuarial frameworks while satisfying governance and model risk requirements.

One of the primary challenges in this study stems from the nature of real-world operational insurance data. Claim records are inherently noisy, heterogeneous, and weakly standardized, requiring substantial effort to understand their structure and idiosyncrasies, including widespread abbreviations, misspellings, informal expressions, and inconsistent narrative styles. Despite extensive data preprocessing, residual data quality issues persist. Many claims contain formally well-structured text that does not reflect true corrective actions, such as adjuster notes, administrative comments, or incomplete diagnostic descriptions. These cases are difficult to detect using simple rules or large-scale automated filters, and no reliable batch solution currently exists to fully eliminate such ambiguities. These data quality challenges directly motivate a promising direction for future research. We plan to deploy the current system in practice and incorporate practitioners in the loop to further address data quality issues, support targeted curation, and enable online learning.

\clearpage

\newpage
\printbibliography

\clearpage

\appendix

\section{Summary of Notation}

\renewcommand{\arraystretch}{1.2}
\begin{longtable}{l p{0.7\linewidth}}
\hline
\textbf{Symbol} & \textbf{Description} \\
\hline
\endfirsthead

\multicolumn{2}{l}{\textbf{Appendix A (continued)}} \\
\hline
\textbf{Symbol} & \textbf{Description} \\
\hline
\endhead

\hline
\endfoot

$n$ & Input token length \\
\( \boldsymbol{x} = (x_1, x_2, \ldots, x_n) \) & Input token sequence (after tokenization) \\
$T$ & Output token length \\
\( \boldsymbol{y} = (y_1, y_2, \ldots, y_T) \) & Output token sequence \\
\( \boldsymbol{y}_{<t} = (y_1, \ldots, y_{t-1}) \) & Previously generated output tokens until $t$ \\
$ \hat{\boldsymbol{y}}_{t} $ & Predicted token distribution at step $t$ \\
$y \in \mathcal{U}^*$ & Reference text represented as a raw Unicode string. \\

$\hat{y} \in \mathcal{U}^*$ & Model-predicted text represented as a raw Unicode string. \\
$N$ & Size of the dataset $\mathcal{D}$ \\
\( \mathcal{D} = \{ (\boldsymbol{x}^{(i)}, y^{(i)}) \}_{i=1}^N \) & Observed dataset \\
$\mathcal{L}$ & Objective function (e.g., the negative log-likelihood) \\
$ \boldsymbol{\theta} $ & Model parameters (frozen or trainable) \\
$\mathcal{U}$ & Set of Unicode characters (\textit{UTF-8} encoded) \\
$\mathcal{U}^*$    & Set of all finite-length Unicode strings (raw text space) \\
$ \mathcal{V} $ & Vocabulary set (e.g., from SentencePiece) \\
$\mathcal{V}^*$    & Set of all finite-length token sequences over $\mathcal{V}$ \\
$ x_{\text{text}} \in \mathcal{U}^* $ & Raw input string (Unicode text) \\
$\mathcal{T}: \mathcal{U}^* \rightarrow \mathcal{V}^*$ & Tokenizer mapping from text to subword tokens \\
$ E \in \mathbb{R}^{|\mathcal{V}| \times d} $ & Token embedding matrix (maps tokens to vectors) \\
$ d $ & Hidden or embedding dimension \\
$ \bm{h}_s \in \mathbb{R}^d $ & Vector representation of the $s$-th token across the model \\
$ \bm{H} \in \mathbb{R}^{n \times d} $ & Sequence matrix representation; each row represents a token vector representation \\ 
$ \bm{h}^{(0)}_s \in \mathbb{R}^d $ & Embedding vector of the $s$-th initial input token \\
$ \bm{\tilde{h}}_s^{(0)} \in \mathbb{C}^{d/2} $ & Complex-valued embedding vector of the $s$-th token \\
$ \bm{\tilde{h}}_{s,k}^{(0,\text{rot})} $ & $k$-th complex component after RoPE rotation at position $s$ \\
$ \boldsymbol{\vartheta} \in \mathbb{R}^{d/2} $ & Angular frequency vector in RoPE \\
$ \omega \in \mathbb{R} $ & Base frequency scaling constant in RoPE, defined as $\omega = 10000^{-1/d}$ \\
$ \boldsymbol{h}_s^{(0,\text{RoPE})} \in \mathbb{R}^d $ & Real-valued embedding vector after RoPE transformation \\
$Q \in \mathbb{R}^{n \times d}$ & Query matrix in self-attention (sequence-level) \\
$K \in \mathbb{R}^{n \times d}$ & Key matrix in self-attention (sequence-level) \\
$V \in \mathbb{R}^{n \times d}$ & Value matrix in self-attention (sequence-level) \\
$ \bm{h}^{(l)}_{\text{in}} $ & Single-token input embedding at layer $l$ \\
$ \bm{h}^{(l)}_{\text{mid}} $ & Single-token embedding after self attention, before FFN, at layer $l$ \\
$ \bm{h}^{(l)}_{\text{out}} $ & Single-token output embedding at layer $l$ \\
$ \bm{H}^{(l)}_{\text{in}} $ & Sequence embedding matrix input to layer $l$ \\
$ \bm{H}^{(l)}_{\text{mid}} $ & Sequence embedding after self attention, before FFN, at layer $l$ \\
$ \bm{H}^{(l)}_{\text{out}} $ & Sequence embedding matrix output from layer $l$ \\
$ \mathcal{F}^{(l)}(\cdot) $ & Transformation function of the $l$-th block (e.g., attention or FFN) \\
$ \operatorname{RMSNorm}(\boldsymbol{v}) $ & Root Mean Square Layer Normalization applied to vector $\boldsymbol{v} \in \mathbb{R}^d$ \\
$\text{Norm}_{\text{in}}^{(l)}$ & Output of RMSNorm applied to $\bm{H}_{\text{in}}^{(l)}$ at layer $l$ \\
$\text{Norm}_{\text{mid}}^{(l)}$ & Output of RMSNorm applied to $\bm{H}_{\text{mid}}^{(l)}$ at layer $l$ \\
$ \boldsymbol{\gamma} \in \mathbb{R}^d $ & Learned scaling vector in RMSNorm \\
$ \epsilon \in \mathbb{R} $ & Small constant added for numerical stability \\
$m_{head}$ & Number of heads in multi-
head causal self-attention \\
$ d_k = d / m_{head} $ & Head dimension, here $d$ is hidden dimension \\
$Q_i \in \mathbb{R}^{n \times d_k}$ & Query matrix of head $i$ in self-attention \\
$K_i \in \mathbb{R}^{n \times d_k}$ & Key matrix of head $i$ in self-attention \\
$V_i \in \mathbb{R}^{n \times d_k}$ & Value matrix of head $i$ in self-attention \\
$ M \in \mathbb{R}^{n \times n} $ & Causal attention mask \\
$W_{i,Q}$ & Query projection matrix of head $i$ in self-attention \\
$W_{i,K}$ & Key projection matrix of head $i$ in self-attention \\
$W_{i,V}$ & Value projection matrix of head $i$ in self-attention \\
$W_O$ & Output projection matrix in self-attention \\
$W_{\text{gate}}$ & Gating projection matrix in FFN (SwiGLU) \\
$W_{\text{up}}$ & Up-projection matrix in FFN (SwiGLU) \\
$W_{\text{down}}$ & Down-projection matrix in FFN (SwiGLU) \\
$ \text{SwiGLU}(a, b) = \text{SiLU}(a) \odot b$ & Activation function in FFN \\
$ W_{\text{out}} \in \mathbb{R}^{|\mathcal{V}| \times d} $ & Output projection matrix mapping hidden state to vocabulary logits \\
$ \mathbf{b}_{\text{out}} \in \mathbb{R}^{|\mathcal{V}|} $ & Output bias vector \\
$ \beta $ & Sampling temperature in softmax \\
$ r $ & LoRA rank (low-rank dimension) \\
$A \in \mathbb{R}^{r \times d_{\text{in}}}$ & LoRA input projection matrix\\
$B \in \mathbb{R}^{d_{\text{out}} \times r}$ & LoRA output projection matrix\\
$W_{\text{frozen}}$ & Frozen base weight matrix \\
$ \Delta W = BA $ & LoRA trainable offset (low-rank update) \\
$d_{\text{in}}$ & Input dimension of the LoRA adapter\\
$d_{\text{out}}$ & Output dimension of the LoRA adapter\\
$ \alpha $ & LoRA scaling factor \\
$\boldsymbol{\theta}_{\text{LoRA}}$ & Trainable parameters introduced by LoRA adapters\\
$ \mathcal{L}_{\text{MLE}}, \mathcal{L}_{\text{LoRA}} $ & Training loss functions \\
$\boldsymbol{z} = (z_1,\ldots,z_L)$ & Concatenated autoregressive training sequence formed by instruction and response tokens \\

$L$ & Total length of the concatenated sequence, $L = n + T$ \\

$\boldsymbol{m} \in \{0,1\}^L$ & Binary supervision mask for token-level loss selection \\

$m_t$ & Mask indicator at position $t$ ($m_t = 1$ if the token contributes to the loss, $0$ otherwise) \\

$\mathcal{L}_{\text{mask}}(\boldsymbol{\theta}_{\text{LoRA}})$ & Masked negative log-likelihood training objective \\

$P_{\text{data}}(y|\boldsymbol{x})$ & Real-world operational conditional distribution \\
$P_{\text{web}}(\boldsymbol{x},y)$  & Pretraining web-scale data distribution \\
$P_{\text{inst}}(\boldsymbol{x},y)$  & Instruction-tuning data distribution \\
$\boldsymbol{\theta}_{\text{pre}}$  &  Parameters after pretraining \\
$\boldsymbol{\theta}_{\text{ft}}$  &  Parameters after fine-tuning \\
$A \subseteq \mathcal{Y}$ & Domain-critical event subset of output space \\
$P_\theta(A|\boldsymbol{x})$     &    Event-level probability induced by the model \\
$|y|$ & String length operator (number of characters) \\
$\mathrm{ED}(\hat{y}, y)$ & Levenshtein edit distance \\
$\mathrm{Sim}_{\mathrm{ED}}(\hat{y}, y)$ & Normalized edit-distance similarity \\
$G_n(y)$ & Multiset of $n$-grams extracted from string $y$ \\
$g$ & An $n$-gram token \\
$c_s(g)$ & Count of $n$-gram $g$ in string $s$ \\
$\mathrm{Prec}_n(\hat{y}, y)$ & $n$-gram precision \\
$p_n(\hat{y}, y)$ & Clipped $n$-gram precision \\
$\mathrm{BLEU}(\hat{y}, y)$ & Sentence-level BLEU score \\
$\mathrm{Sim}_{\mathrm{tfidf}}(\hat{y}, y)$ & TF--IDF cosine similarity \\
$\mathbf{v}(y)$ & TF--IDF vector representation of string $y$ \\
$\|\cdot\|_2$ & Euclidean norm \\
$\mathrm{Sim}_{\mathrm{BERT}}(\hat{y}, y)$ & BERT sentence embedding cosine similarity \\
$\mathrm{Sim}_{\text{Human}}(\cdot,\cdot)$ & score assigned by human annotators following the scoring guideline \\
$\mathbf{e}(y)$ & Sentence embedding of string $y$ \\
$P_{\mathrm{BS}}(\hat{y}, y)$ & BERTScore precision \\
$R_{\mathrm{BS}}(\hat{y}, y)$ & BERTScore recall \\
$\mathrm{BERTScore}_{\mathrm{F1}}(\hat{y}, y)$ & BERTScore F1 \\
$f_{\mathrm{BLEURT}}$ & BLEURT neural scoring function \\
$\mathrm{BLEURT}(\hat{y}, y)$ & BLEURT semantic similarity score \\
$g_{\phi}$ & Frozen large language model used as an evaluator \\
$\phi$ & Fixed parameters of the evaluator LLM \\
$\pi(\hat{y}, y)$ & Deterministic prompting function mapping a prediction--reference pair to an evaluation prompt \\
$\mathrm{Judge}(\hat{y}, y)$ & LLM-as-a-Judge evaluation score \\
$\mathrm{Parse}(\cdot)$ & Function extracting a scalar score from the LLM response \\
$\kappa$ & High-quality decision threshold selected via Youden's $J$ statistic ($\kappa=0.77$). \\

\end{longtable}

\section{Implementation and Reproducibility} \label{appendix:implementation}

The training system is implemented in PyTorch \citep{paszke2019pytorch} which serves as the core deep-learning framework underlying all model construction and optimization. PyTorch provides automatic differentiation and large-scale numerical computation support, enabling efficient gradient-based training of neural networks and scalable execution on modern GPU hardware.

Our training pipeline is built on HuggingFace Transformers, \citep{wolf2020transformers}\footnote{\url{https://huggingface.co/docs/transformers}} which offers standardized implementations of modern LLMs together with their associated tokenizers and text-generation utilities. This framework allows us to load pretrained language models, perform fine-tuning, and conduct evaluation within a unified and reproducible interface, without re-implementing low-level model components.

To adapt large pretrained models efficiently, we employ the PEFT, \citep{mangrulkar2022peft}\footnote{\url{https://github.com/huggingface/peft}} library for parameter-efficient fine-tuning. Instead of updating all model parameters, PEFT inserts a small number of trainable modules into selected layers while keeping the original backbone weights frozen. This substantially reduces GPU memory usage and computational cost, enabling controlled experiments on large-scale language models without full retraining.

Training is orchestrated using TRL's SFTTrainer, \citep{vonwerra2023trl}\footnote{\url{https://github.com/huggingface/trl}}, a high-level supervised fine-tuning framework designed for instruction-style learning and structured-output generation. SFTTrainer manages the full training loop, including data loading, batching, optimization, checkpointing, and evaluation, providing a reliable abstraction for large-model fine-tuning while allowing task-specific customization.

To support the learning objective defined in Section~\ref{subsec:task-setup-masked-training}, we implement a custom token-level loss function with fine-grained masking over the generated sequence. Gradients are applied only to semantically meaningful target spans, such as corrective-action fields, while prompt tokens and non-supervised regions are excluded from optimization. This design enforces structural controllability and prevents spurious learning signals from irrelevant tokens.

All experimental runs are parameterized through a unified configuration file specifying dataset paths, preprocessing options, LoRA injection settings, optimizer and scheduler parameters, random seeds, and hardware configurations. This enables systematic ablation studies and ensures full experimental reproducibility across multiple runs.

The training pipeline supports distributed execution under \texttt{torch.distributed}, allowing scalable multi-GPU training on high-performance computing clusters. It further includes automatic experiment directory management, periodic checkpointing, and centralized logging via Weights \& Biases \footnote{\url{https://wandb.ai}} for real-time monitoring of training dynamics and model behavior. Upon completion, trained LoRA adapters and held-out test sets are automatically exported to support controlled downstream benchmarking and error analysis.

\section{Formal Definitions of Automated Evaluation Metrics} \label{appendix:Metrics}

In this section, we formally define all automated evaluation metrics used in section~\ref{sec:eval}. Let $y$ denote the reference corrective action and $\hat{y}$ the model-predicted action, both represented as natural language strings. All metrics quantify the similarity between $\hat{y}$ and $y$ from different perspectives.

\paragraph{Edit distance similarity.}

Let $\mathrm{ED}(\hat{y},y)$ denote the character-level Levenshtein edit distance between $\hat{y}$ and $y$. We define the normalized edit-distance similarity as
\begin{equation}
\mathrm{Sim}_{\mathrm{ED}}(\hat{y},y)
= 1 - \frac{\mathrm{ED}(\hat{y},y)}{\max\left(\lvert \hat{y} \rvert, \lvert y \rvert\right)},
\end{equation}
where $\lvert \cdot \rvert$ denotes string length (number of characters).

\paragraph{N-gram precision.}
Let $\mathcal{G}_n(s)$ denote the multiset of n-grams (tokens) extracted from a string $s$, and let $c_s(g)$ denote the count of n-gram $g$ in $s$. We define n-gram precision as

\begin{equation}
\mathrm{Prec}_n(\hat{y},y)
= \frac{\sum_{g \in \mathcal{G}_n(\hat{y})} \min\!\big(c_{\hat{y}}(g),\,c_{y}(g)\big)}
{\sum_{g \in \mathcal{G}_n(\hat{y})} c_{\hat{y}}(g)}.
\end{equation}


\paragraph{Sentence BLEU.}
We adopt sentence-level BLEU as implemented in sacreBLEU \citep{post2018call}, based on the original BLEU formulation \citep{papineni2002bleu}.
The clipped $n$-gram precision is defined as
\begin{equation}
p_n(\hat{y}, y)
= \frac{\sum_{g \in \mathcal{G}_n(\hat{y})}
\min\!\big(c_{\hat{y}}(g),\, c_{y}(g)\big)}
{\sum_{g \in \mathcal{G}_n(\hat{y})} c_{\hat{y}}(g)}.
\end{equation}

Sentence BLEU is computed as the geometric mean of modified $n$-gram precisions with a brevity penalty:
\begin{equation}
\mathrm{BLEU}(\hat{y},y)
= \mathrm{BP} \cdot \exp\!\left( \sum_{n=1}^{N} w_n \log p_n(\hat{y},y) \right),
\end{equation}
where $w_n$ are uniform weights, $N=4$ in our experiments, and the brevity penalty is
\begin{equation}
\mathrm{BP} =
\begin{cases}
1, & |\hat{y}| > |y|, \\
\exp\!\left(1 - \frac{|y|}{|\hat{y}|}\right), & |\hat{y}| \le |y|.
\end{cases}
\end{equation}

\paragraph{TF--IDF cosine similarity.}
Let $\mathcal{D} = \{y_1,\dots,y_T,\hat{y}_1,\dots,\hat{y}_T\}$ denote the combined corpus consisting of all reference and predicted actions. A TF--IDF vectorizer is fitted on $\mathcal{D}$ with $n$-gram range $(1,2)$. For any string $s$, let $\mathbf{v}(s) \in \mathbb{R}^d$ denote its TF--IDF representation under this shared vocabulary and inverse document frequency weighting.

The TF--IDF cosine similarity between a prediction $\hat{y}$ and its reference $y$ is defined as
\begin{equation}
\mathrm{Sim}_{\mathrm{tfidf}}(\hat{y},y)
= \frac{\mathbf{v}(\hat{y})^\top \mathbf{v}(y)}
{\|\mathbf{v}(\hat{y})\|_2 \, \|\mathbf{v}(y)\|_2}.
\end{equation}

\paragraph{BERT cosine similarity.}
We adopt a pretrained sentence embedding model from the Sentence-Transformers framework. Specifically, we use \texttt{all-mpnet-base-v2} \citep{song2020mpnet} to encode each string into a fixed-dimensional vector.

Let $\mathbf{e}(s) \in \mathbb{R}^d$ denote the sentence embedding of a string $s$ produced by the encoder. The BERT cosine similarity between a prediction $\hat{y}$ and its reference $y$ is defined as
\begin{equation}
\mathrm{Sim}_{\mathrm{BERT}}(\hat{y},y)
= \frac{\mathbf{e}(\hat{y})^\top \mathbf{e}(y)}
{\|\mathbf{e}(\hat{y})\|_2 \, \|\mathbf{e}(y)\|_2}.
\end{equation}

\paragraph{BERTScore (F1).}
We use BERTScore as a token-level semantic similarity metric based on contextualized embeddings. Given a prediction $\hat{y}$ and a reference $y$, both are first tokenized and encoded by a pretrained transformer encoder into contextual token embeddings.

Let $\mathbf{h}_i(\hat{y}) \in \mathbb{R}^d$ denote the embedding of the $i$-th token in $\hat{y}$, and $\mathbf{h}_j(y) \in \mathbb{R}^d$ the embedding of the $j$-th token in $y$. The pairwise token similarity is defined by cosine similarity
\[
s_{ij} = \frac{\mathbf{h}_i(\hat{y})^\top \mathbf{h}_j(y)}
{\|\mathbf{h}_i(\hat{y})\|_2 \, \|\mathbf{h}_j(y)\|_2}.
\]

BERTScore precision and recall are defined as
\begin{equation}
P_{\mathrm{BS}}(\hat{y},y) = \frac{1}{|\hat{y}|} \sum_{i} \max_{j} s_{ij}, 
\qquad
R_{\mathrm{BS}}(\hat{y},y) = \frac{1}{|y|} \sum_{j} \max_{i} s_{ij}.
\end{equation}

The BERTScore F1 is given by
\begin{equation}
\mathrm{BERTScore}_{\mathrm{F1}}(\hat{y},y)
= \frac{2 P_{\mathrm{BS}}(\hat{y},y) R_{\mathrm{BS}}(\hat{y},y)}
{P_{\mathrm{BS}}(\hat{y},y) + R_{\mathrm{BS}}(\hat{y},y)}.
\end{equation}.

\paragraph{BLEURT.}
We adopt BLEURT as a learned neural evaluation model. Formally, let
\[
f_{\theta_{\mathrm{BLEURT}}} : \mathcal{U}^* \times \mathcal{U}^* \rightarrow \mathbb{R}
\]
denote a pretrained neural scoring function that maps a reference action $y$ and a predicted action $\hat{y}$ to a real-valued semantic similarity score.

The BLEURT score is defined as
\begin{equation}
\mathrm{BLEURT}(\hat{y}, y) = f_{\theta_{\mathrm{BLEURT}}}(y, \hat{y}),
\end{equation}
where $f_{\theta_{\mathrm{BLEURT}}}$ is instantiated as a Transformer-based regression model fine-tuned to predict human judgment scores over text pairs. In our experiments, we use the publicly released checkpoint \texttt{Elron/bleurt-base-128} hosted on the HuggingFace Model Hub and loaded via the \texttt{transformers} library \citep{wolf2020transformers}. Given a pair $(y, \hat{y})$, the model outputs a single scalar logit, which is used directly as the BLEURT score.

\paragraph{LLM-as-a-Judge score.}
We implement an LLM-based evaluator to score predicted actions according to a predetermined evaluation rubric.

Formally, let
\[
g_{\phi} : \mathcal{X} \rightarrow \mathcal{T}
\]
denote a pretrained LLM with fixed parameters $\phi$, and let
\[
\pi : \mathcal{U}^* \times \mathcal{U}^* \rightarrow \mathcal{X}
\]
be a deterministic prompting function that maps a prediction--reference pair $(\hat{y}, y)$ into a structured evaluation prompt.

The LLM-as-a-Judge score is defined as
\begin{equation}
\mathrm{Judge}(\hat{y}, y)
= \mathrm{Parse}\!\left(g_{\phi}(\pi(\hat{y}, y))\right),
\end{equation}
where $g_{\phi}(\pi(\hat{y}, y))$ denotes the model-generated evaluation response, and $\mathrm{Parse}(\cdot)$ extracts a scalar score according to a fixed scoring evaluation rubric. In our experiments, $g_{\phi}$ is instantiated as \texttt{ChatGPT-4o-mini}, the scoring rubric is defined in Table \ref{tab:human-guideline}, and all evaluations are performed using deterministic decoding.

\section{Prompt Engineering}

For prompt engineering base models (DeepSeek-R1 and Qwen-Instruct), we use a fixed instruction prompt that constrains the base model to output only the corrective action from a warranty claim description. The prompt enforces a single-line output with no explanation, reasoning, or auxiliary text, ensuring direct comparability with structured ground-truth labels.

\begin{lstlisting}
You are given a warranty claim description.

Your task:
Output ONLY the corrective action.

Rules:
- Do NOT explain.
- Do NOT include any reasoning.
- Do NOT include any extra text.
- Do NOT include formatting symbols.
- Output exactly one line.

Output format:
corrective action include: <action>

Claim description:
{claim_text}
\end{lstlisting}

\section{Equivalence Between Maximum Likelihood and Conditional KL Minimization}
\label{appendix:kl proof}

Let $\boldsymbol{x}$ denote the input variable, $y$ the corresponding output variable,
$P_{\mathrm{data}}(y\mid \boldsymbol{x})$ the data-generating conditional distribution,
and $P(y\mid \boldsymbol{x};\boldsymbol{\theta})$ a parametric model.

\begin{align*}
\boldsymbol{\theta}^\star
&= \arg\min_{\boldsymbol{\theta}}
\mathbb{E}_{\boldsymbol{x}\sim P_{\mathrm{data}}(\boldsymbol{x})}
\Big[
\mathrm{KL}\!\left(
P_{\mathrm{data}}(y\mid \boldsymbol{x})
\,\|\, 
P(y\mid \boldsymbol{x};\boldsymbol{\theta})
\right)
\Big] \\[6pt]
&= \arg\min_{\boldsymbol{\theta}}
\mathbb{E}_{\boldsymbol{x}\sim P_{\mathrm{data}}(\boldsymbol{x})}
\mathbb{E}_{y\sim P_{\mathrm{data}}(y\mid \boldsymbol{x})}
\left[
\log P_{\mathrm{data}}(y\mid \boldsymbol{x})
-
\log P(y\mid \boldsymbol{x};\boldsymbol{\theta})
\right]
\quad \text{(definition of KL)} \\[6pt]
&= \arg\min_{\boldsymbol{\theta}}
\left(
\mathbb{E}_{(\boldsymbol{x},y)\sim P_{\mathrm{data}}}
\big[\log P_{\mathrm{data}}(y\mid \boldsymbol{x})\big]
-
\mathbb{E}_{(\boldsymbol{x},y)\sim P_{\mathrm{data}}}
\big[\log P(y\mid \boldsymbol{x};\boldsymbol{\theta})\big]
\right)
\quad \text{(rearranging expectations)} \\[6pt]
&= \arg\min_{\boldsymbol{\theta}}
\left(
-\mathbb{E}_{(\boldsymbol{x},y)\sim P_{\mathrm{data}}}
\big[\log P(y\mid \boldsymbol{x};\boldsymbol{\theta})\big]
\right)
\quad \text{(dropping terms independent of $\boldsymbol{\theta}$)} .
\end{align*}

This establishes the equivalence between conditional KL minimization and
maximum likelihood estimation under the data distribution.

\section{Examples of Format Compliance and Validity} \label{appendix:format}

\paragraph{Definition} We define the output schema $\mathcal{S}$ as a single-line corrective-action expression of the form
\[
\mathcal{S} \;:=\; \texttt{corrective action(s) include(d): } \; \langle \text{action} \rangle
\]

In our dataset $\mathcal{D}$, all expected outputs follow this schema. Accordingly, the prompts provided to the model explicitly require generated outputs to conform to the same format.

\textbf{Non-Compliant Output Example:} 

\textit{DeepSeek-R1 + Prompt Output:} </technician> So, the problem is that there's a nail embedded in the side wall of the left rear tire. The technician needs to determine the correct action to fix this. So, the correct action is to replace the left rear tire with

\textbf{Format Compliant Output Example:}

\textit{DeepSeek-R1+ Fine-tune Output:} Corrective actions included: repaired right front wheel

\textbf{Output without actionable content(non-Valid) Example:}

\textit{DeepSeek-R1 Output:} loose belt" condition So, the problem is that the engine's drive belt is loose, causing a rattle when the engine starts. The possible causes are: 1. Belt tensioner pulley worn 2. Belt tensioner spring

\textbf{Output with actionable content(Valid) Example:}

\textit{DeepSeek-R1 + Prompt Output:} </technician> So, the issue is that the ball joints are loose or damaged. The customer was told that the ball joints needed to be replaced. So, the corrective action is to \textbf{replace the ball joints}. Wait, but the problem is that

\section{Benchmark Comparison with Commercial Large Language Models} \label{appendix:Commercial}

We provide a detailed empirical comparison between our fine-tuned model and five SoTA commercial LLMs: Gemini-2.5-Flash, Claude Haiku 4.5, GPT-4o-mini, GPT-4.1, and GPT-5.2. All benchmark evaluations reflect the state of the market as of February 2026, when these final experiments were completed. 

We evaluate semantic alignment with ground truth using BERT cosine similarity on both the full valid set and the HQ subset. We first report summary statistics, including standard summary statistics and the proportion of outputs exceeding a high-quality threshold, to quantify relative performance across models. We then present distributional comparisons in the subsequent figures to illustrate how these differences manifest across the full range of similarity scores.

Our comparative analysis reveals that among commercial LLMs, Gemini-2.5-Flash delivers the strongest performance. However, it still falls short of our fine-tuned 8B parameter model, despite the latter's significantly smaller scale compared to industry-leading commercial architectures. Notably, within the OpenAI suite, GPT-4o-mini outperformed more advanced or costly models such as GPT-4.1 and GPT-5.2. This provides compelling evidence that raw model scale and SoTA status do not necessarily translate to superiority in specialized insurance tasks, where domain-specific training proves to be the decisive factor.

\begin{table}[htbp]
\centering
\caption{Summary statistics of BERT cosine similarity of DeepSeek-R1 + Fine-tune and SoTA commercial LLM on the full evaluation set ($N=1500$).}
\label{tab:bert_summary_full}
\begin{tabular}{lrrrrrr}
\toprule
Model & Count & Mean & Std. & Median & Max & $p(\mathrm{Sim}_{\mathrm{BERT}} \ge \kappa)$ \\
\midrule
DeepSeek-R1 + Fine-tune         & 1500 & 0.869 & 0.159 & 0.929 & 1.000 & 0.791 \\
Gemini-2.5-Flash  & 1500 & 0.799 & 0.160 & 0.857 & 0.977 & 0.685 \\
Claude-Haiku-4.5  & 1500 & 0.757 & 0.170 & 0.803 & 0.974 & 0.575 \\
GPT-4o-mini       & 1500 & 0.787 & 0.163 & 0.844 & 0.974 & 0.656 \\
GPT-4.1           & 1500 & 0.749 & 0.163 & 0.793 & 0.977 & 0.561 \\
GPT-5.2           & 1500 & 0.719 & 0.167 & 0.762 & 1.000 & 0.472 \\
\bottomrule
\end{tabular}

\vspace{4pt}
\footnotesize{\emph{Note.} $\kappa$ denotes the correct decision threshold selected via Youden’s $J$ statistic to best approximate human classification; here $\kappa=0.77$.}
\end{table}

\begin{table}[htbp]
\centering
\caption{Summary statistics of BERT cosine similarity DeepSeek-R1 + Fine-tune and SoTA commercial LLM on the HQ subset ($N=1334$).}
\label{tab:bert_summary_hq}
\begin{tabular}{lrrrrrr}
\toprule
Model & Count & Mean & Std. & Median & Max & $p(\mathrm{Sim}_{\mathrm{BERT}} \ge \kappa)$ \\
\midrule
DeepSeek-R1 + Fine-tune        & 1334 & 0.899 & 0.132 & 0.946 & 1.000 & 0.863 \\
Gemini-2.5-Flash  & 1334 & 0.817 & 0.146 & 0.867 & 0.977 & 0.733 \\
Claude-Haiku-4.5  & 1334 & 0.776 & 0.159 & 0.819 & 0.970 & 0.624 \\
GPT-4o-mini       & 1334 & 0.806 & 0.149 & 0.856 & 0.974 & 0.705 \\
GPT-4.1           & 1334 & 0.767 & 0.152 & 0.809 & 0.977 & 0.605 \\
GPT-5.2           & 1334 & 0.736 & 0.155 & 0.773 & 1.000 & 0.507 \\
\bottomrule
\end{tabular}

\vspace{4pt}
\footnotesize{\emph{Note.} $\kappa$ denotes the correct decision threshold selected via Youden’s $J$ statistic to best approximate human classification; here $\kappa=0.77$.}
\end{table}

\begin{figure}[htbp]
    \centering
    \includegraphics[width=0.95\textwidth]{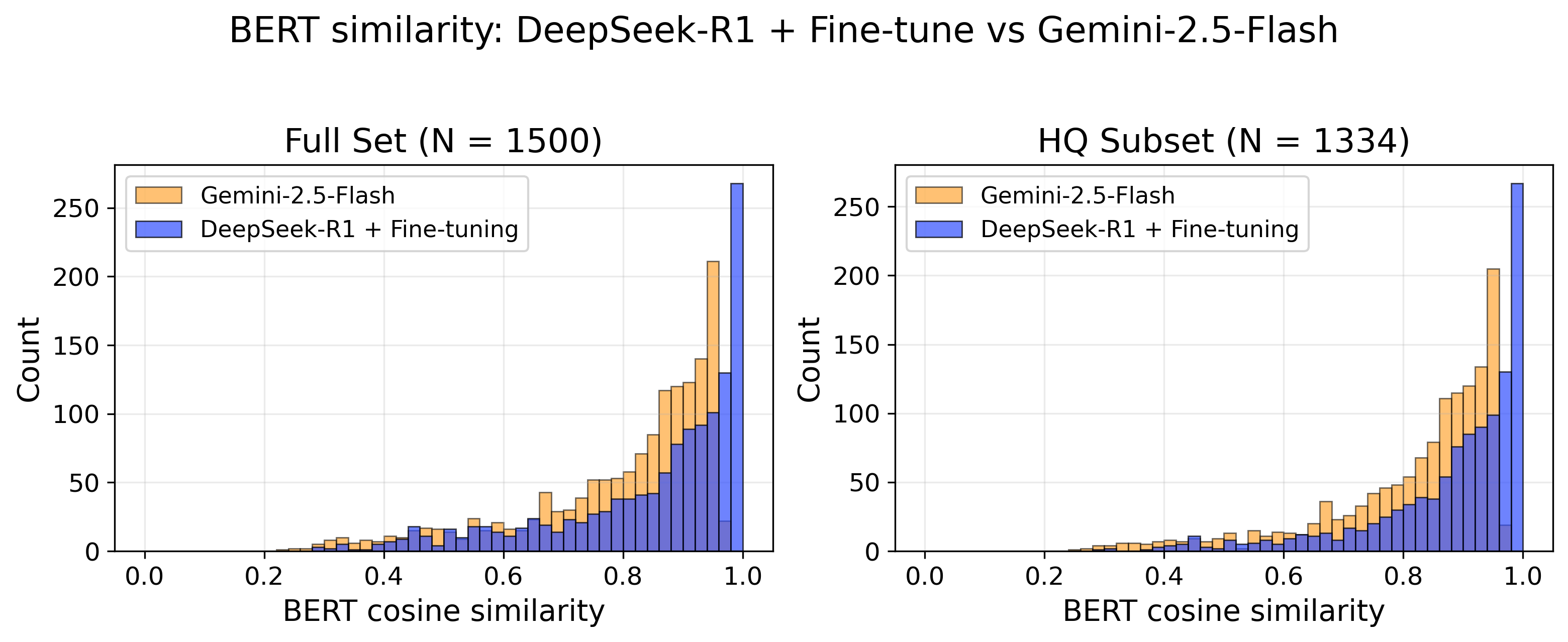}
    \caption{BERT cosine similarity: DeepSeek-R1 + Fine-tune versus Gemini-2.5-Flash on the full evaluation set and HQ subset.}
    \label{fig:vs_gemini}
\end{figure}

\begin{figure}[htbp]
    \centering
    \includegraphics[width=0.95\textwidth]{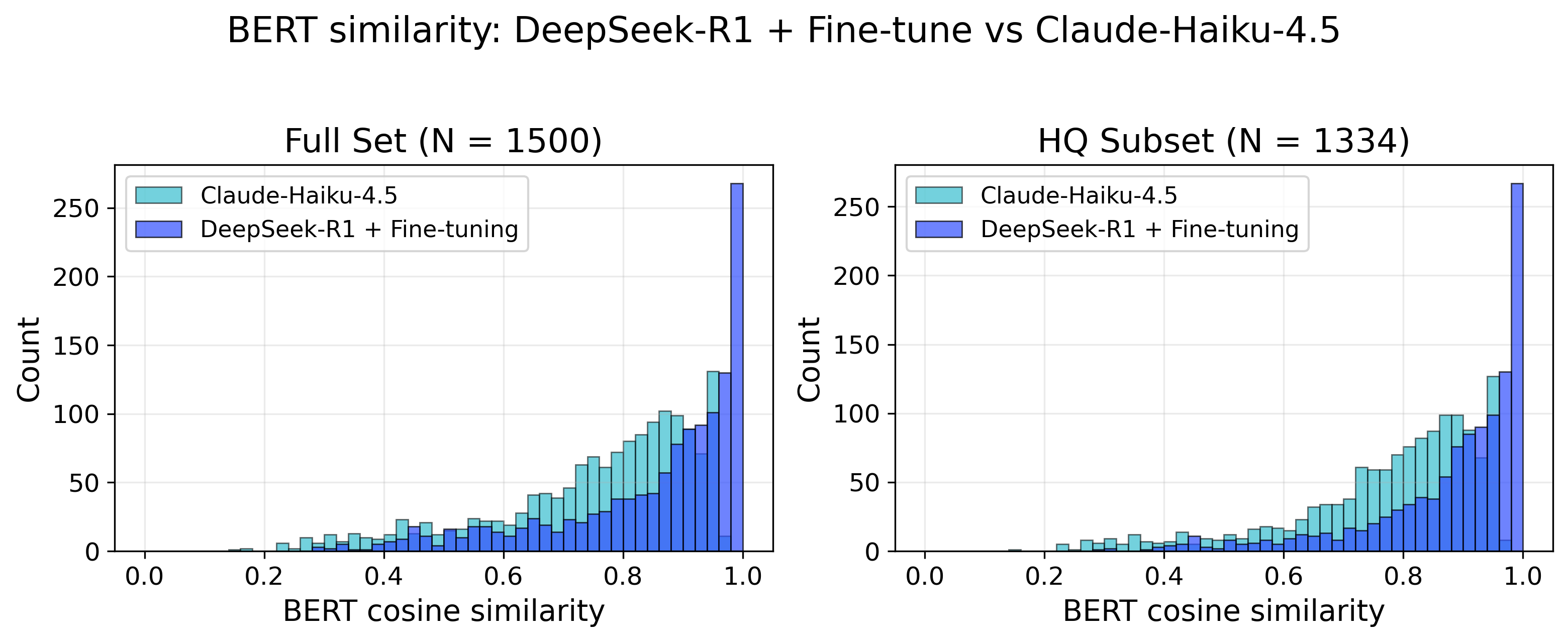}
    \caption{BERT cosine similarity: DeepSeek-R1 + Fine-tune versus Claude Haiku 4.5 on the full evaluation set and HQ subset.}
    \label{fig:vs_claude}
\end{figure}

\begin{figure}[htbp]
    \centering
    \includegraphics[width=0.95\textwidth]{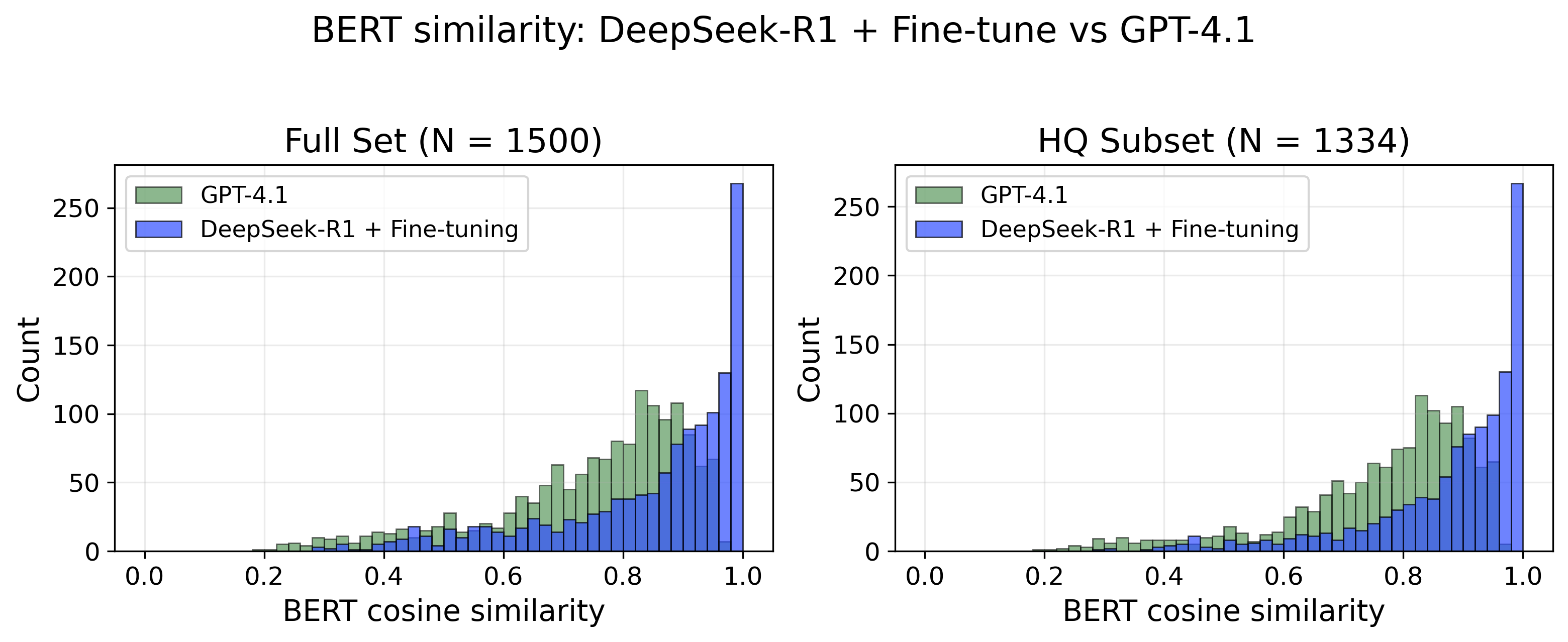}
    \caption{BERT cosine similarity: DeepSeek-R1 + Fine-tune model versus GPT-4.1 on the full evaluation set and HQ subset.}
    \label{fig:vs_gpt4.1}
\end{figure}

\begin{figure}[htbp]
    \centering
    \includegraphics[width=0.95\textwidth]{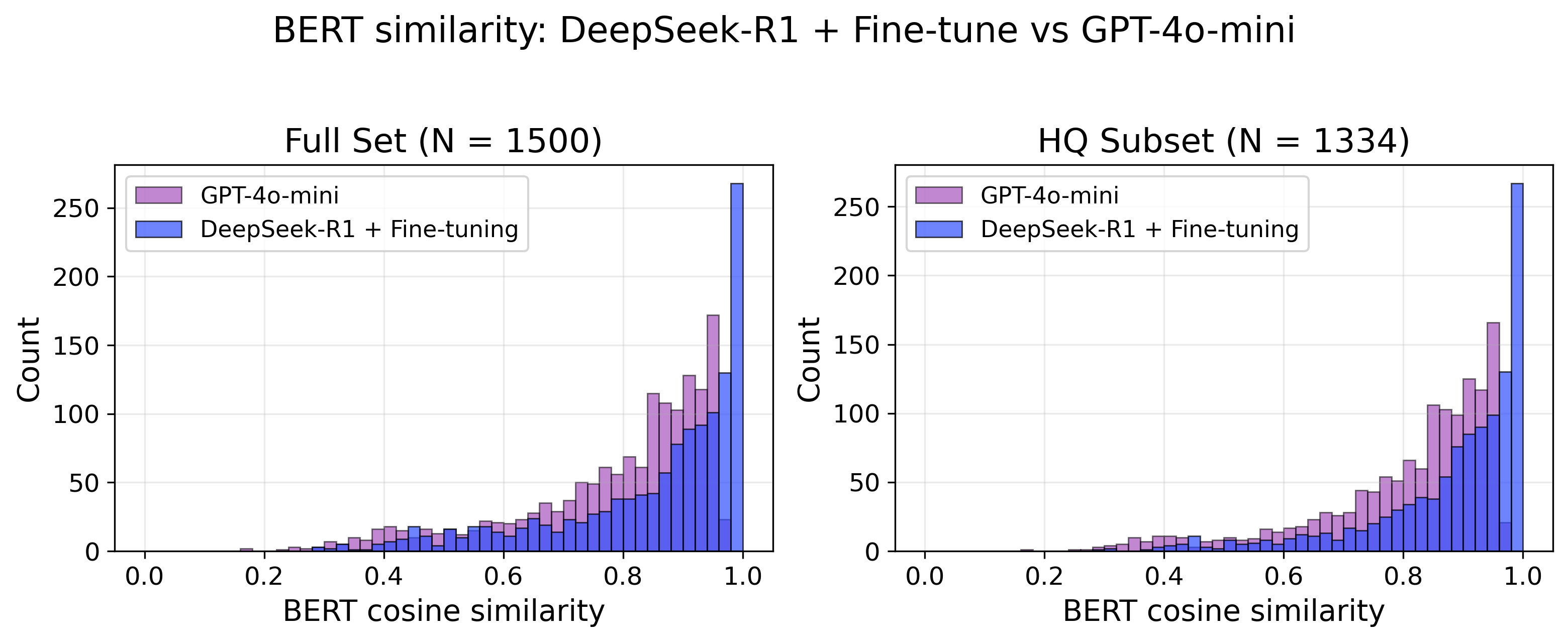}
    \caption{BERT cosine similarity: DeepSeek-R1 + Fine-tune model versus GPT-4o-mini on the full evaluation set and HQ subset.}
    \label{fig:vs_gpt4o_mini}
\end{figure}

\begin{figure}[htbp]
    \centering
    \includegraphics[width=0.95\textwidth]{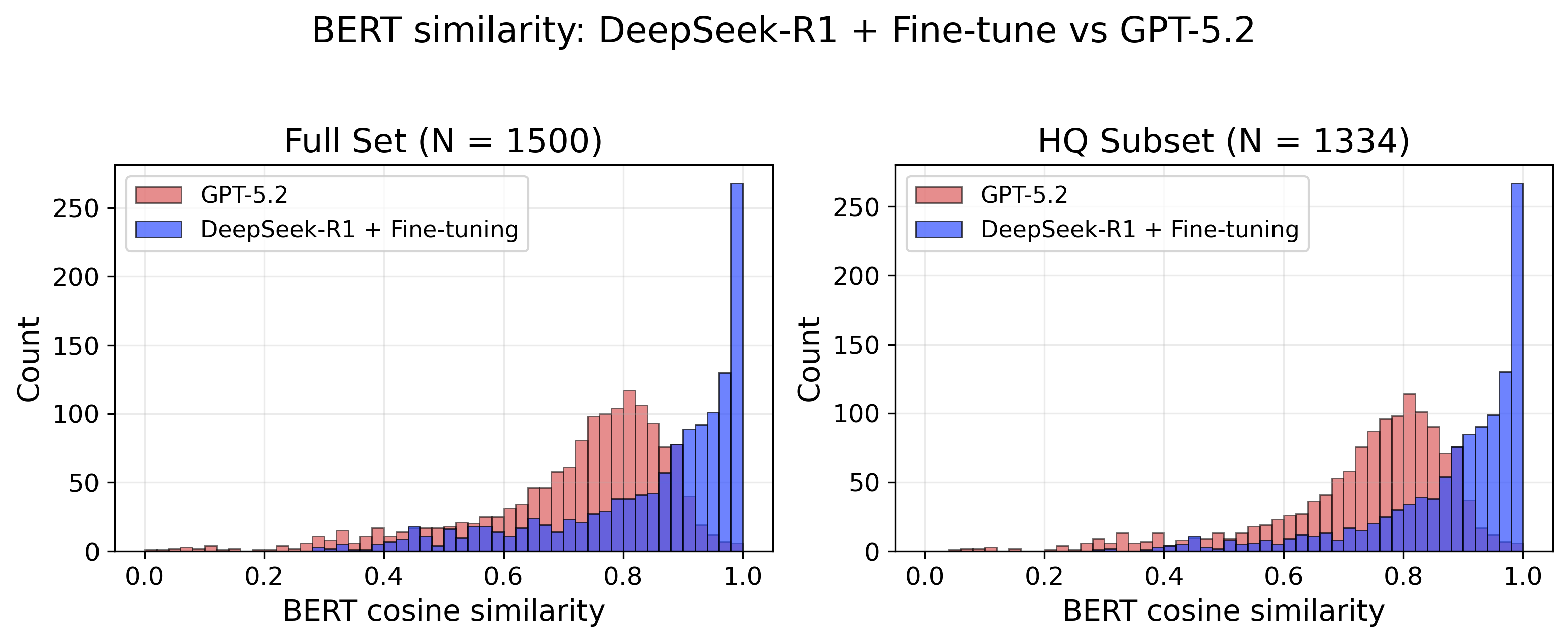}
    \caption{BERT cosine similarity: DeepSeek-R1 + Fine-tune versus GPT-5.2 on the full valid set and HQ subset.}
    \label{fig:vs_gpt5.2}
\end{figure}

\section{Case Study: Repair vs. Replace in Non-Repairable Tire Claims}
\label{sec:tire_case_study}

According to \citet{tire2008}, punctures located on the tire sidewall are generally considered non-repairable and require tire replacement rather than repair. This rule is well established in automotive safety standards and is routinely enforced in warranty claim operations.

This practice is clearly reflected in our dataset. Among the $2,953$ tire-related claims that contain lexical indicators of sidewall damage (e.g., ``inside wall,'' ``in side wall''), only $4$ cases resulted repair-type corrective actions (e.g., ``repair tire,'' ``patch tire,'', or ``remove nail''), yielding an empirical repair probability of  $\frac{4}{2953} \approx 1.36 \times 10^{-3}$. This extreme imbalance closely mirrors real-world repair decisions for sidewall damage. If we assume the repair probability follows a Binomial distribution (not suitable for the imbalance case, again introducing conservative bias), then the expected number of repairs is $4$. The fine-tuned model predicts only $1$ repair-type actions among the $2,953$ tire-related claims, whereas non-fine-tuned models predict $262$ (DeepSeek-R1 + Prompt) and $65$ (Qwen-Instruct + Prompt), respectively. These large deviations indicate severe, statistically significant distributional shifts, with non-fine-tuned models systematically overestimating repair probability for practically non-repairable sidewall damage claims.

In contrast, the fine-tuned model's predictions are statistically indistinguishable from the empirical reference distribution. Human evaluation suggests that non-fine-tuned models often rely on generic automotive narratives (e.g., ``punctures can often be patched''), reflecting common consumer-level knowledge rather than real-world industry practice. The fine-tuned model, trained on historical warranty claim outcomes, internalizes this highly asymmetric conditional distribution, aligning its prediction with real-world practice.

\section{Error Categories and Representative Examples (81 cases)}
\label{appendix:Error categoried}

\subsection{Category I: Surface Form Divergence with Semantic Equivalence (19 cases, 23.4\%)}

The first category consists of cases in which the model output is judged to be semantically equivalent to the reference but differs in surface form.

\textbf{Example.}  
\textit{Expected Output:} Corrective actions included: replace remove and replace tire due to impact to the tire and for riding on while flat.  

\textit{Model Prediction:} Corrective actions included: replace tire.

In such cases, the model produces a concise and operationally correct corrective action, while the reference output embeds explanatory or repetitive phrases. From a practical standpoint, the model output is considered clearer and more suitable for downstream automation. The low evaluation score in these cases primarily reflects limitations of automated evaluation metrics rather than deficiencies in predictive accuracy.

\subsection{Category II: Target Granularity Mismatch (16 cases, 19.8\%)}

This category includes cases where the model captures the intended repair but expresses it at a different level of component granularity than the reference. Such mismatches arise, for example, when the model outputs a higher-level assembly while the reference specifies a subcomponent, or vice versa.

\textbf{Example.}  
\textit{Expected Output:} Corrective actions included: replace panoramic roof track and shade assembly.  

\textit{Model Prediction:} Corrective actions included: replace sunroof assembly.

Although both outputs refer to the same general repair operation, the reference adopts a more fine-grained specification. In practice, predictions of this form remain informative, but the difference in granularity can introduce ambiguity for downstream use, particularly in labor time estimation and parts costing. As a result, this class of errors reflects a meaningful deviation in resolution, even when the underlying repair direction is aligned.

\subsection{Category III: Repair Omission (17 cases, 21.0\%)}

This category includes cases where the reference contains multiple corrective actions, but the model predicts only a subset, omitting one or more required repairs.

\textbf{Example.}  
\textit{Expected Output:} Corrective actions included: remove and replace heater hose assembly; remove and replace engine assembly.  

\textit{Model Prediction:} Corrective actions included: remove and replace engine.

In these cases, the model identifies a major repair operation but fails to capture the full set of required actions. Such omissions can lead to systematic underestimation of total repair scope and associated costs. From an actuarial perspective, partial predictions of this form are particularly concerning, as they may distort severity assessment, reserve estimates, and downstream pricing assumptions.

\subsection{Category IV: Incorrect Repair Action (6 cases, 7.4\%)}

In this class of errors, the model correctly identifies the affected component but predicts an incorrect repair action.

\textbf{Example.}  
\textit{Expected Output:} Corrective actions included: reseal timing cover.  

\textit{Model Prediction:} Corrective actions included: replace timing cover.

Although the repair target is correctly identified, the distinction between resealing and replacement corresponds to fundamentally different repair procedures, with materially different implications for labor hours, parts costs, and overall claim severity. Errors of this type, while relatively infrequent, are operationally significant, as they alter the cost structure of the predicted repair and can lead to substantial distortion in projected claim costs.

\subsection{Category V: Incorrect Repair Part (23 cases, 28.4\%)}

This final and most severe category consists of cases in which the model predicts an incorrect repair component.

\textbf{Example.}  
\textit{Expected Output:} Corrective actions included: remove and replace oil filter adapter.  

\textit{Model Prediction:} Corrective actions included: remove and replace oil cooler.

In these cases, the model fails to identify the correct repair target, resulting in outputs that are semantically misaligned with the underlying repair requirement. Such errors fundamentally compromise the accuracy of the prediction and pose the highest risk to downstream automation, including cost estimation, claim triaging, and workflow integration. Consistent with their severity, this category accounts for the largest share of genuinely harmful failures observed in our analysis.

\clearpage

\end{document}